\documentclass[10pt,twocolumn,letterpaper]{article}

\usepackage[pagenumbers]{iccv} 

\definecolor{iccvblue}{rgb}{0.21,0.49,0.74}
\usepackage[pagebackref,breaklinks,colorlinks,allcolors=iccvblue]{hyperref}
\usepackage[accsupp]{axessibility}
\usepackage{multirow} 
\usepackage{color, colortbl}

\def\paperID{5776} 
\def\confName{ICCV}
\def\confYear{2025}

\title{Fair Generation without Unfair Distortions: \\Debiasing Text-to-Image Generation with Entanglement-Free Attention}

\author{Jeonghoon Park*\\
KAIST\\
{\tt\small jeonghoon\_park@kaist.ac.kr}
\and
Juyoung Lee*\\
Kakao Corp.\\
{\tt\small zealota11@gmail.com}
\and
Chaeyeon Chung\\
KAIST\\
{\tt\small cy\_chung@kaist.ac.kr}
\and
Jaeseong Lee\\
Yonsei University\\
{\tt\small jasonlee1995@yonsei.ac.kr}
\and
Jaegul Choo\\
KAIST\\
{\tt\small jchoo@kaist.ac.kr}
\and
Jindong Gu\\
University of Oxford\\
{\tt\small jindong.gu@eng.ox.ac.uk}
}

\begin{document}

\twocolumn[{
\maketitle
\begin{center}
    \vspace{-5mm}
    \centering 
    \includegraphics[width=\textwidth]{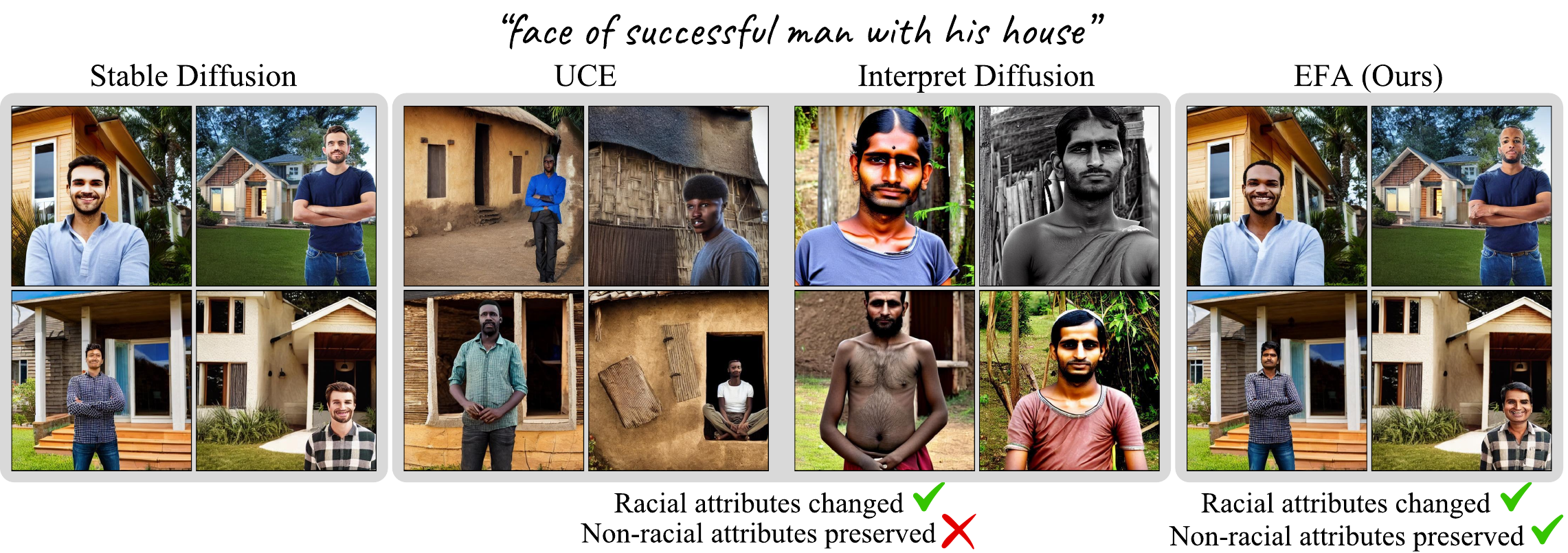}
    \captionof{figure}{
    Generated images from the pretrained model (\eg, Stable Diffusion~\cite{2022stable}), by previous debiasing methods (\eg, UCE~\cite{gandikota2024unified} and Interpret Diffusion~\cite{li2024self}), and ours.
    Previous debiasing methods often alter the non-target attributes (\eg, house) while modifying target bias attributes (\eg, racial attributes). In contrast, our approach preserves non-target attributes during bias mitigation (\ie, changing the racial attributes without modifying the house in the background), thereby preventing the introduction of unintended biases. The corresponding aligned images were generated using the same random seed.
    }
\label{fig:teaser}
\end{center}
}]

\begin{abstract}
Recent advancements in diffusion-based text-to-image (T2I) models have enabled the generation of high-quality and photorealistic images from text. However, they often exhibit societal biases related to gender, race, and socioeconomic status, thereby potentially reinforcing harmful stereotypes and shaping public perception in unintended ways. While existing bias mitigation methods demonstrate effectiveness, they often encounter attribute entanglement, where adjustments to attributes relevant to the bias (\ie, target attributes) unintentionally alter attributes unassociated with the bias (\ie, non-target attributes), causing undesirable distribution shifts. To address this challenge, we introduce Entanglement-Free Attention (EFA), a method that accurately incorporates target attributes (\eg, White, Black, and Asian) while preserving non-target attributes (\eg, background) during bias mitigation. At inference time, EFA randomly samples a target attribute with equal probability and adjusts the cross-attention in selected layers to incorporate the sampled attribute, achieving a fair distribution of target attributes. Extensive experiments demonstrate that EFA outperforms existing methods in mitigating bias while preserving non-target attributes, thereby maintaining the original model’s output distribution and generative capacity.
\end{abstract}
    
\section{Introduction}

Diffusion-based text-to-image (T2I) models~\cite{nichol2021glide, rombach2022high, ramesh2022hierarchical, saharia2022photorealistic, betker2023improving} have gained significant attention due to their ability to reflect user intent through text input, making them highly applicable across various domains.
However, despite their advancements, these T2I models often perpetuate socially sensitive biases and stereotypes such as gender, race, or socioeconomic status~\cite{cho2023dalleval, bianchi2023easily, friedrich2023FairDiffusion}.
Such societal biases can reinforce discriminatory perceptions of certain social groups and pose ethical and practical risks in real-world environments, where biased outputs can influence public perception and decision making.

Previous works attempt to mitigate such biases via two primary approaches. 
One widely explored strategy is finetuning, where pretrained models are adjusted using additional training on curated datasets~\cite{orgad2023editing, gandikota2024unified, shen2024finetuning}. 
Despite its effectiveness, this method often entails substantial computational costs~\cite{shen2024finetuning}.
Also, it may result in a decline in image quality and diversity due to the relatively small size of the finetuning dataset compared to the original pretraining dataset~\cite{schuhmann2022laion}.
Since this approach inherently modifies the parameters of the model, it may lead to unintended changes in the output distribution.
An alternative line of research explores inference-time steering, which avoids modifying the pretrained model and instead applies adjustments only during inference~\cite{friedrich2023FairDiffusion, chuang2023debiasVL, kim2023stereotyping, Shrestha2024FairRAG, Parihar2024balancingAct, li2024self}.
Although this approach effectively reduces bias while preserving the parameters of the original model, we observe that it still introduces unintended alterations in the generated outputs.

Fig.~\ref{fig:teaser} illustrates this issue by presenting images generated using the prompt ``face of successful man with his house".
The pretrained model correctly interprets the prompt by generating images of a man with a house but predominantly produces individuals from a limited racial background.
When existing bias mitigation methods are applied, the representation of underrepresented racial attributes, such as Black and Indian individuals, is successfully improved.
However, they also introduce unintended shifts in non-racial attributes.
Specifically, they alter the material and architectural style of the house, associating certain racial groups with impoverished or deteriorated living conditions, thus reinforcing misleading associations between social groups and socioeconomic status.
This finding suggests that previous methods may involve attribute entanglement between attributes relevant to the target bias (\ie, target attribute, such as racial attributes) and attributes unassociated with the target bias (\ie, non-target attribute, such as the appearance of the house), potentially leading to unintended alterations in the non-target attributes during the bias mitigation process.
Without careful disentanglement, bias mitigation can unintentionally alter contextual details, degrade output diversity, and even introduce new biases. 

To address this issue, we propose Entanglement-Free Attention (EFA), a method that accurately incorporates target attributes while preserving non-target attributes. 
In this paper, we focus on human-centric bias, which plays a crucial role in shaping societal perceptions. 
Our EFA operates by leveraging predefined target attributes (\eg, White, Black, Asian, and Indian) that contribute to bias (\eg, race). 
To guide the model to focus on semantically relevant regions (e.g., face and body), EFA is trained to enhance target attributes in regions that exhibit high semantic relevance to the counterfactual prompts (\ie, prompts with alternative attributes other than the target), while human segmentation masks further constrain its effect to human regions. 
By employing both target and counterfactual prompts in training, EFA learns to dynamically adjust the enhancement strength based on the presence of the target attribute, applying stronger modifications when it is underrepresented.
During inference, to ensure a fair distribution of target attributes in the generated output, one attribute is randomly sampled with equal probability. 
EFA then enhances the corresponding features in bias-relevant regions without relying on human-provided masks, effectively disentangling target and non-target attributes while preserving the original model’s generative capacity and contextual consistency.

In summary, our key contributions are as follows:
\begin{itemize}
    \item  We identify the issue of attribute entanglement, where bias mitigation unintentionally alters non-target attributes, leading to unintended distribution shifts in diffusion-based T2I models.
    
    \item We propose EFA, a novel debiasing approach that decouples target and non-target attributes, ensuring that bias mitigation is applied only to the intended target attributes while preserving the integrity of non-target attributes.
    
    \item Quantitative and qualitative experiments demonstrate that our approach successfully mitigates bias while maximally preserving the generation capability of the original model, outperforming previous methods.

\end{itemize}

\section{Related works}
\begin{figure*}[t!]
\vspace{-5mm}
\centering
\includegraphics[width=\textwidth]{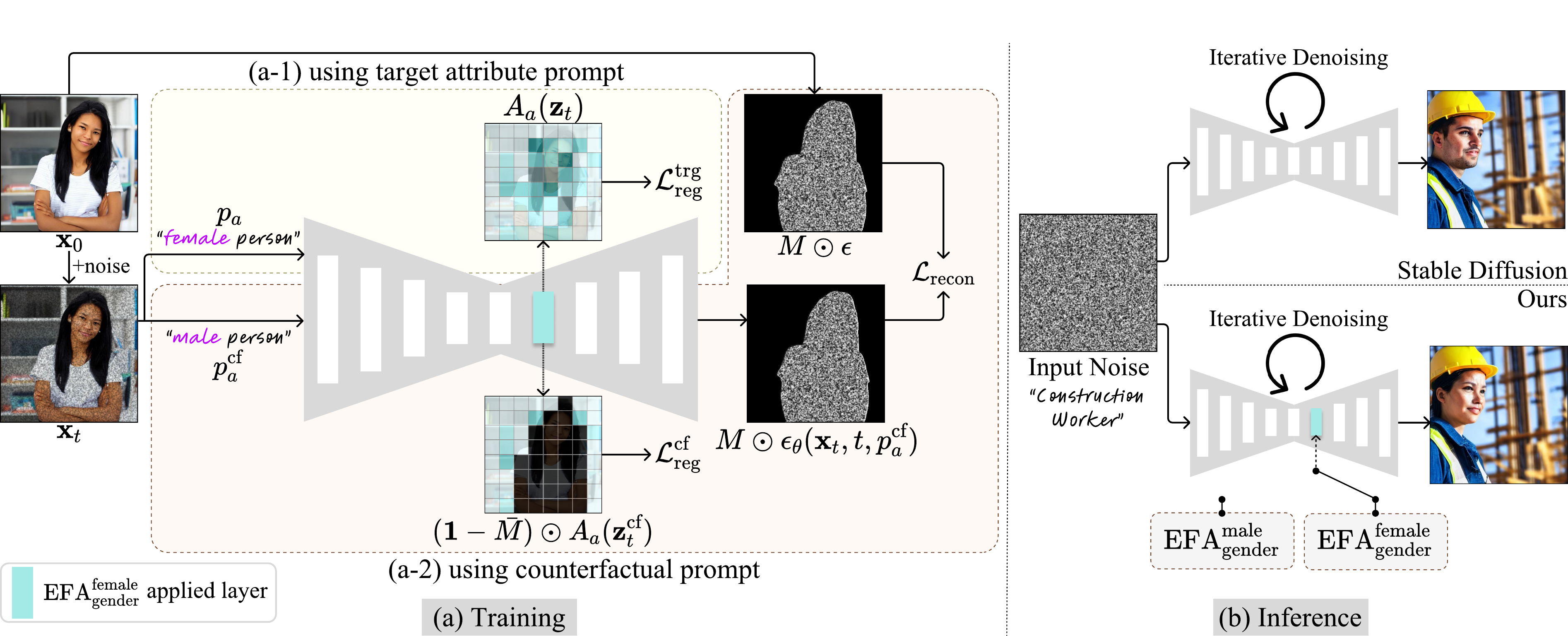}
\caption{
Overview of our method. (a): EFA is applied to selected layers and trained to reconstruct the image $\mathbf{x}_0$ (a-1) when given $p_a$, the prompt containing the target attribute $a$, or (a-2) its counterfactual prompt, $p_a^\text{cf}$.
We leverage a human segmentation mask $M$ to minimize the effect of EFA on the attention map of non-target attribute regions.
(b): During the inference, our method generates outputs with a fair distribution of target attributes using the EFA of the randomly sampled target attribute.
}
\vspace{-2mm}
\label{fig:method}
\end{figure*}
Existing studies on mitigating bias in T2I generative models have mainly followed two approaches.
The first approach involves directly modifying the parameters of diffusion-based T2I models. 
For instance, \citet{orgad2023editing} and \citet{gandikota2024unified} proposed a method to adjust the weights of the cross-attention module to alleviate the bias.
Similarly, \citet{shen2024finetuning} finetune the text encoder to alleviate face-related biases (\eg, gender, race).
However, since finetuning is conducted on a significantly smaller dataset compared to the pretraining dataset, such approach may potentially degrade original output quality or diversity. 
Furthermore, the approach of \citet{shen2024finetuning} requires substantial computational costs for training the backbone model.

Another line of research has explored inference-time steering approaches, where the pretrained model remains unchanged while adjustments are applied during inference. 
\citet{friedrich2023FairDiffusion} proposed a bias mitigation framework that employs semantic guidance~\cite{brack2023sega} to manipulate the target attributes by referring to registered information of a concept (\eg, nurse) and its bias (\eg, gender). 
However, this cannot mitigate bias for unregistered concepts, limiting its practicality.
\citet{chuang2023debiasVL} have developed a method to remove bias from text embeddings, and \citet{kim2023stereotyping} trains the soft prompts for mitigating bias.
Beyond employing the text embeddings, \citet{Shrestha2024FairRAG} introduced a technique that projects a reference image into text-compatible tokens, allowing the model to leverage the features of external images, such as gender or race for image generation. 
Additionally, \citet{Parihar2024balancingAct} and \citet{li2024self} proposed methods to manipulate the h-space~\cite{kwon2023diffusion} of the model during inference.
While these methods aim to mitigate bias without directly altering the backbone parameters, they often struggle to preserve the original context of the generated images.

Recently, \citet{zhou2024association} addressed association-engendered stereotypes arising from co-occurrence of multiple objects in an image. 
They rely on a CLIP model trained to predict biases from prompts, requiring prior knowledge of the biases to be mitigated. 
In contrast, our approach aims to address subtle biases induced by contextual variations beyond the predefined association bias definitions.
For instance, if a particular social group is predominantly depicted in low-resource work environments, while another is shown in corporate office settings, such context variations can reinforce cognitive biases, subtly implying disparities in professional status.
Furthermore, it is impractical to anticipate all possible forms of such bias in advance.

\section{Entanglement-free attention}

\subsection{Overview}
Without loss of generality, we assume a given target bias $C$ (\eg, gender) along with its corresponding set of target attributes $\mathcal{A}_C=\{a_1, a_2, \ldots, a_n\}$ (\eg, $\{\text{`female'}, \text{`male'}\}$) whose elements should be uniformly distributed in the outputs.
Then, our approach aims to alleviate the bias $C$ by adjusting the distribution of target attributes in $\mathcal{A}_C$ while preserving the non-target attributes such as scene context and surrounding objects.

To achieve this, we introduce EFA, which leverages the principle of cross-attention mechanism that assigns desired features to appropriate locations.
Each target attribute $a_i \in \mathcal{A}_C$ is handled by a dedicated EFA module (\ie, $\text{EFA}_C^{a_i}$), ensuring faithful representation of $a_i$ in the outputs while minimizing its influence on non-target attribute regions.
During inference, to enforce a fair distribution of target attributes in the generated output, we randomly sample an attribute $a_i$ from $\mathcal{A}_C$ with equal probability and incorporate its representation using the corresponding EFA (Fig.~\ref{fig:method} (b)).

To enhance the fidelity of $a_i$, $\text{EFA}_C^{a_i}$ is trained to reconstruct images depicting $a_i$ when its representation is subtle by introducing counterfactual attributes $a_i^\text{cf} \in \mathcal{A}_C\setminus \{a_i\}$ through counterfactual prompts.
EFA is guided to enhance the target attribute in areas exhibiting strong semantic relevance to the counterfactual prompt through the cross-attention module.
Additionally, human segmentation masks are used during training to constrain EFA’s influence to the human region.
This ensures that $\text{EFA}_C^{a_i}$ enhances the fidelity of $a_i$ while preventing unintended modifications to non-target attributes.

\subsection{Architecture of EFA}
The original cross-attention module takes an intermediate spatial feature $\textbf{z}_t$, which is derived by from a noised image $\textbf{x}_t$ and a prompt $p$ using the denoising model $\phi$ (\ie, $\textbf{z}_t = \phi(\textbf{x}_t, p)$.
$\textbf{z}_t$ is transformed into a query representation using the mapping layer $W_q$.
The text prompt embeddings $\pi(p)$ are projected into key and value representations through $W_k$ and $W_v$, respectively.
The output of the cross-attention layer is formulated as:
\begin{equation*}
    \text{Cross-attention}(\mathbf{z}_t)=\text{softmax}(\frac{QK^\top}{\sqrt{d}})V,
\end{equation*}
where $d$ indicates the dimension of the query and key vectors, and $Q=\textbf{z}_tW_q$, $K=\pi(p)W_k$, and  $V=\pi(p)W_v$ denote the query, key, and value matrices, respectively.

\begin{figure}[t!]
\centering
\includegraphics[width=\columnwidth]{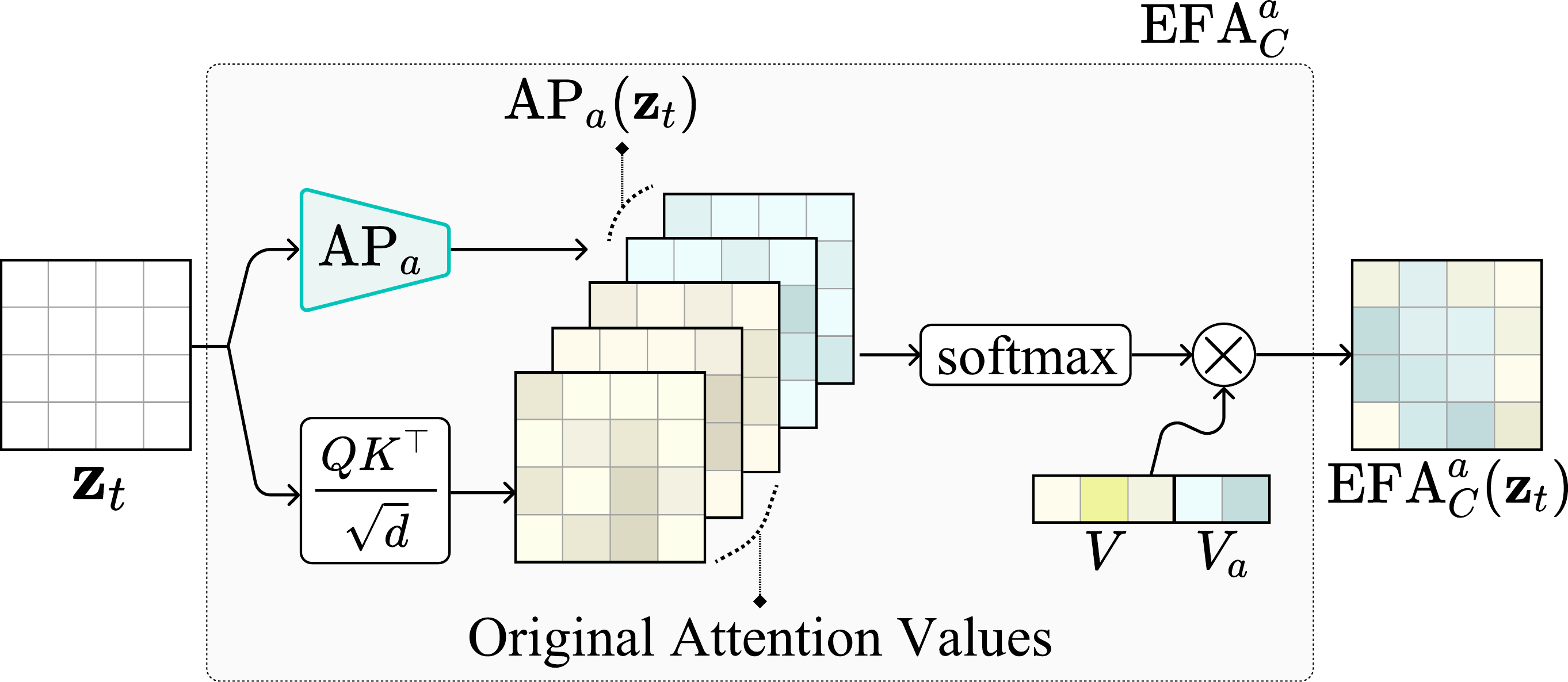}
\caption{
Architecture of EFA. We denote $a_i\in \mathcal{A}_C$ as $a$ in this figure for notational simplicity.
}
\vspace{-4.5mm}
\label{fig:architecture}
\end{figure}

As illustrated in Fig.~\ref{fig:architecture}, EFA modifies the cross-attention to integrate the desired attribute $a_i\in \mathcal{A}_C$ into the output while preserving the original attention structure.
Specifically, EFA takes the intermediate feature $\textbf{z}_t$ and predicts the additional attention values that determine where and to what extent $a_i$ should be enhanced.
This is achieved using the attention value predictor (AP), a lightweight module composed of three convolutional layers and two SiLU~\cite{silu} activation functions.
Further details of the architecture are provided in the Supplementary.

The predicted attention value (\ie, $\text{AP}_{a_i}(\mathbf{z}_t)$) are concatenated with the original cross-attention's attention values (\ie, $\frac{QK^\top}{\sqrt{d}}$), and the softmax function is applied.
After applying softmax, the original value vectors from the user prompt (\ie, $V$) and the value vectors of $a_i$ are weighted accordingly, as in the standard cross-attention operation.
Formally, EFA incorporates $a_i$ into the cross-attention as follows:
\begin{equation*}
    \text{EFA}_C^{a_i}(\mathbf{z}_t)=\text{softmax}([\frac{QK^\top}{\sqrt{d}},\text{AP}_{a_i}(\mathbf{z}_t)])[V, V_{a_i}],
\end{equation*}
where $V_{a_i}=\pi(p_{a_i})W_v$, $p_{a_i}$ indicates the text prompt representing ${a_i}$, and $[\cdot, \cdot]$ indicates the concatenation.

When building an $\text{EFA}_C$ for a target bias $C$, all AP modules share the same convolutional backbone while maintaining distinct output channels designated for each attribute in $\mathcal{A}_C$.
This design allows AP models to be effectively trained together, ensuring that EFAs of target attributes for a given bias are jointly learned while avoiding the inefficiency of training multiple independent APs.

\subsection{Training}

\noindent\textbf{Dataset construction.}
To train $\text{EFA}_C^{a_i}$ to represent a target attribute $a_i$ (\eg, female) while preserving non-target attributes, we create a dataset $\mathcal{D}_{a_i}$ that captures $a_i$ and provides pseudo supervision on the human regions that contain $a_i$.
Specifically, we generate images using the original model with prompts explicitly describing $a_i$ (\eg, ``a female person"). 
For each image, we extract a human segmentation mask $M$ using Grounded SAM2~\cite{ravi2024sam2segmentimages, liu2023grounding}, assigning 1 to human regions and 0 elsewhere. 
Since this study focuses on human-centric bias, we utilize a human segmentation mask to serve as the basis for regulating the influence of $a_i$ during training.
The resulting $\mathcal{D}_{a_i}$ consists of pairs of images and their segmentation masks.

\noindent\textbf{Training EFA.}
At inference, the extent of the features that need to be adjusted by EFA to faithfully reflect $a_i$ may vary across inputs.
For instance, with the prompt ``nurse", intermediate features may exhibit a high level of femininity, requiring minimal adjustment by $\text{EFA}_\text{gender}^\text{female}$.
In contrast, intermediate features of the prompt ``doctor" may have stronger masculinity, requiring enhanced representation of female by $\text{EFA}_\text{gender}^\text{female}$ to incorporate `female' into the output.
To handle this variability, $\text{EFA}_C^{a_i}$ is trained in two different scenarios
: (1) when intermediate features contain the attribute $a_i$ (\eg, female), and (2) when intermediate features contain its counterfactual attribute $a_i^\text{cf} \in \mathcal{A}_C \setminus \{a_i\}$ (\eg, male).
We denote $a_i$ as $a$ throughout this section for simplicity.

Let $\mathbf{x}_0$ be an image and $M$ be its corresponding segmentation mask, sampled from $\mathcal{D}_{a}$, where $a$ denotes a target attribute (\eg, female).
In the first scenario, where the intermediate features already encode $a$, $\text{EFA}_C^a$ does not need to enhance $a$ but should instead preserve the original cross-attention output to maintain the non-target attributes.
To enforce this behavior, we provide the model with a paired input: a prompt $p_a$ explicitly specifying $a$ (\eg, ``female person") and a noised image $\mathbf{x}_t$ that inherently contains $a$, as depicted in Fig.~\ref{fig:method} (a-1).

To prevent unnecessary alterations, we introduce an L1 regularization term that encourages the attention values associated with $V_a$ (\eg, attention values associated with the value vectors of ``female") to be zero.
This regularization loss is defined as:
\begin{equation*}
    \mathcal{L}_{\text{reg}}(\mathbf{z}_t,W) = {\lVert W\odot A_a(\mathbf{z}_t)\rVert_1},
\end{equation*}
where $\mathbf{z}_t=\phi(\textbf{x}_t, p_a)$, $W$ is a spatial mask used to constrain the effect of the loss, and $\odot$ represents the element-wise multiplication. 
Also, $A_a(\mathbf{z}_t)$ denotes the attention values (after softmax) associated with the value vectors $V_a$ given $\mathbf{z}_t$.
More precisely, when $A(\mathbf{z}_t)=\text{softmax}([\frac{QK^\top}{\sqrt{d}},\text{AP}_a(\mathbf{z}_t)])$, we define $A_a(\mathbf{z}_t)$ as the submatrix of $A(\mathbf{z}_t)$ obtained by selecting values indexed by $I_{V_a}$, where $I_{V_a}$ denotes the set of indices corresponding to $V_a$.

In this scenario, we set $W=\mathbf{1}$, a mask of ones, to apply the loss over all spatial positions.
This regularization loss minimizes unnecessary modifications to the original cross-attention output introduced by $\text{EFA}_C^a$.
Hereafter, we denote $\mathcal{L}_{\text{reg}}(\mathbf{z}_t,\mathbf{1})$ as $\mathcal{L}_{\text{reg}}^\text{trg}$ for brevity.

In the second scenario, the intermediate features contain a counterfactual attribute $a^\text{cf}$ (\eg, male) induced by bias of prompt (\eg, ``photo of a doctor" implicitly favoring male representation).
Here, $\text{EFA}_C^a$ should suppress $a^\text{cf}$ and enhance $a$ (\eg, female) while preserving non-target attributes.
To simulate this, we train $\text{EFA}_C^a$ to counteract counterfactual features introduced during the denoising process of a diffusion-based T2I model, which generates clean images by iteratively denoising from pure noise.

To encourage representation of $a$ in human-related regions, we define a reconstruction loss leveraging the segmentation mask $M$ to constrain its effect on human areas.
The objective is to predict the added noise $\epsilon$ given $\mathbf{x}_t$ and a counterfactual prompt $p_a^\text{cf}$, formulated as follows:
\begin{equation*}
\mathcal{L}_\text{recon} = \mathbb{E}_{\mathbf{x}_0, t, p_{a}^\text{cf}, \epsilon \sim N(0,1)} \left[ \|M \odot \left(\epsilon-\epsilon_{\theta}(\mathbf{x}_t, t, p_{a}^\text{cf})\right)\|^2_2 \right],
\end{equation*}
where $\epsilon_\theta$ denotes the model with $\text{EFA}_C^a$.
Through the reconstruction process, $\text{EFA}_C^a$ is encouraged to enhance $a$ in regions where cross-attention assigns high similarity to $a^\text{cf}$, reflecting their semantic relevance.
Using human masks ensures that $a$ is accurately reconstructed in human regions while mitigating the influence of $a^\text{cf}$.

Although $\mathcal{L}_\text{reg}^\text{trg}$ reduces unintended modifications, it does not explicitly confine its impact on human regions, potentially affecting non-target attributes.
To address this, we impose an additional L1 regularization on the attention values corresponding to $V_a$, applied outside the human region, as follows:
\begin{equation*}
\mathcal{L}_{\text{reg}}^{\text{cf}}=\mathcal{L}_\text{reg}(\mathbf{z}_t^\text{cf}, \mathbf{1}-\bar{M}),
\end{equation*}
where $\mathbf{z}_t^\text{cf}=\phi(\textbf{x}_t, p_a^\text{cf})$, and $\bar{M}$ indicates the binary mask resized to match the resolution of the input feature of $\text{EFA}_C^a$.

Finally, the overall training loss is formulated as:
\begin{equation*}
    \mathcal{L}_\text{tot} = \mathcal{L}_\text{recon} + \mathcal{L}_\text{reg}^*,
\end{equation*}
where $\mathcal{L}_\text{reg}^* = \lambda_1\mathcal{L}_\text{reg}^\text{trg} + \lambda_2\mathcal{L}_\text{reg}^\text{cf} $ and $\lambda_1$ and $\lambda_2$ are hyperparameters to control the relative importance between the losses.
During training, only the parameters of $\text{AP}_a$ in $\text{EFA}_C^a$ are updated.
Note that EFA is applied to the up-block layers with an input resolution of 16×16, as lower-resolution features capture higher-level semantic information.
The choice of applied layers is discussed in Section~\ref{subsec:anal_layer}.
The method of extending EFA to handle multiple biases is provided in the Supplementary.

\subsection{Inference}
To mitigate the bias $C$ and ensure fair generation, we randomly sample an attribute $a_i$ with an equal probability from $\mathcal{A}_C$ and apply the corresponding $\text{EFA}_C^{a_i}$ to the original model.
This approach enables users to easily control the attribute distribution by adjusting the sampling frequency, without additional training.
Note that our method requires no masks at inference and generates debiased images directly from noise, offering a practical and efficient solution.

\definecolor{Gray}{gray}{0.92}

\begin{table*}[t]
\centering
\resizebox{\textwidth}{!}{
\setlength{\tabcolsep}{1.0em}
\def\arraystretch{1.0}%
\begin{tabular}{ l l | c c c c | c  c c c }
\toprule 

\multicolumn{2}{c|}{\multirow{2}{*}{}}

&\multicolumn{4}{c|}{{$\mathcal{T}_\text{basic}$}}  
&\multicolumn{4}{c}{{$\mathcal{T}_\text{complex}$}}  
\\
\cmidrule(lr){3-6} \cmidrule(lr){7-10}
&
& \multicolumn{1}{c}{{Bias}}  
& \multicolumn{3}{c|}{{Non-target attribute P.}}
& \multicolumn{1}{c}{{Bias}}  
& \multicolumn{3}{c}{{Non-target attribute P.}}
\\
\cmidrule(lr){3-3} \cmidrule(lr){4-6}
\cmidrule(lr){7-7} \cmidrule(lr){8-10}
& Method
& DR $\downarrow$
& PSNR $\uparrow$
& LPIPS $\downarrow$
& DINO $\uparrow$
& DR $\downarrow$
& PSNR $\uparrow$
& LPIPS $\downarrow$
& DINO $\uparrow$

\\
\midrule
\parbox[t]{2mm}{\multirow{6}{*}{\rotatebox[origin=c]{90}{Gender}}} &
Original SD 
& 0.71 & - & - & - 
& 0.71 & - & - & - \\

& Concept Algebra~\cite{wang2023concept} 
& 0.59 & 21.10 & 0.1169 & 0.823
& 0.69 & 16.69 & 0.1852 & 0.834 \\

& UCE~\cite{gandikota2024unified}
& 0.34 & 21.04 & 0.1374 & 0.757
& 0.49 & 16.78 & 0.2202 & 0.747 \\

& Finetuning Diffusion~\cite{shen2024finetuning} 
& 0.48 & 22.62 & 0.1166 & 0.814
& 0.56 & 19.27 & 0.1611 & 0.857 \\

& Interpret Diffusion~\cite{li2024self}
& 0.26 & 17.18 & 0.2290 & 0.616
& 0.31 & 14.01 & 0.3211 & 0.618 \\

& \cellcolor{Gray}EFA (Ours) 
& \cellcolor{Gray}\textbf{0.06} & \cellcolor{Gray}\textbf{32.52} & \cellcolor{Gray}\textbf{0.0411} & \cellcolor{Gray}\textbf{0.916}
& \cellcolor{Gray}\textbf{0.06} & \cellcolor{Gray}\textbf{29.70} & \cellcolor{Gray}\textbf{0.0492} & \cellcolor{Gray}\textbf{0.941} \\

\midrule

\parbox[t]{2mm}{\multirow{6}{*}{\rotatebox[origin=c]{90}{Race}}} &

Original SD 
& 0.60 & - & - & - 
& 0.55 & - & - & - \\

& Concept Algebra~\cite{wang2023concept}
&0.64 & 21.47 & 0.1164 & 0.839
&0.58 & 16.62 & 0.1888 & 0.841 \\

& UCE~\cite{gandikota2024unified} 
& 0.27 & 21.55 & 0.1261 & 0.787
& 0.41 & 17.81 & 0.1891 & 0.801\\

& Finetuning Diffusion~\cite{shen2024finetuning}
& 0.29 & 18.96 & 0.1795 & 0.709
& 0.37 & 16.46 & 0.2263 & 0.771\\

& Interpret Diffusion~\cite{li2024self}
& 0.16 & 16.87 & 0.2416 & 0.584
& 0.23 & 13.73 & 0.3338 & 0.584\\

& \cellcolor{Gray}EFA (Ours)
& \cellcolor{Gray}\textbf{0.04} & \cellcolor{Gray}\textbf{30.93} & \cellcolor{Gray}\textbf{0.0353} & \cellcolor{Gray}\textbf{0.938}
& \cellcolor{Gray}\textbf{0.06} & \cellcolor{Gray}\textbf{28.55} & \cellcolor{Gray}\textbf{0.0421} & \cellcolor{Gray}\textbf{0.958}\\

\midrule

\parbox[t]{2mm}{\multirow{6}{*}{\rotatebox[origin=c]{90}{Gender $\times$ Race}}} &
Original SD 
& 0.56 & - & - & - 
& 0.50 & - & - & - \\
& Concept Algebra~\cite{wang2023concept}
& 0.51 & 20.12 & 0.1325 & 0.805
& 0.47 & 15.88 & 0.2031 & 0.818 \\

& UCE~\cite{gandikota2024unified}
& 0.16 & 19.84 & 0.1575 & 0.702
& 0.32 & 16.79 & 0.2198 & 0.746 \\

& Finetuning Diffusion~\cite{shen2024finetuning} 
& 0.24 & 19.97 & 0.1570 & 0.746
& 0.28 & 17.24 & 0.2064 & 0.795 \\

& Interpret Diffusion~\cite{li2024self}
& 0.16 & 16.91 & 0.2449 & 0.466
& 0.20 & 13.66 & 0.3529 & 0.475 \\

& \cellcolor{Gray}EFA (Ours)
& \cellcolor{Gray}\textbf{0.03} & \cellcolor{Gray}\textbf{25.58} & \cellcolor{Gray}\textbf{0.0684} & \cellcolor{Gray}\textbf{0.853}
& \cellcolor{Gray}\textbf{0.05} & \cellcolor{Gray}\textbf{23.78} & \cellcolor{Gray}\textbf{0.0795} & \cellcolor{Gray}\textbf{0.903} \\

\bottomrule
\end{tabular}
}
\caption{
Comparison to baselines. P. is the abbreviation for preservation. The best values are in bold. Our method achieves superior bias mitigation performance while effectively preserving non-target attributes, significantly outperforming previous approaches.
}
\label{tab:recent_comparison}
\end{table*}
\definecolor{Gray}{gray}{0.92}

\begin{table}[t]
\centering
\resizebox{\columnwidth}{!}{
\setlength{\tabcolsep}{1.0em}
\def\arraystretch{1.0}%
\begin{tabular}{ l l | c  c }
\toprule

& Method
& FID $\downarrow$
& CLIP-T $\uparrow$
\\
\midrule
-
&Original SD
&- & 26.17 \\
\midrule

\parbox[t]{2mm}{\multirow{5}{*}{\rotatebox[origin=c]{90}{Gender}}}
& Concept Algebra~\cite{wang2023concept}
&2.41 & 26.02 \\
& UCE~\cite{gandikota2024unified}
&11.65 & 25.17 \\
& Finetuning Diffusion~\cite{shen2024finetuning}
&1.92 & 25.79 \\
& Interpret Diffusion~\cite{li2024self}
&15.78 & 24.80 \\
& \cellcolor{Gray}EFA (Ours) 
& \cellcolor{Gray}\textbf{0.23} & \cellcolor{Gray}\textbf{26.03} \\

\midrule

\parbox[t]{2mm}{\multirow{5}{*}{\rotatebox[origin=c]{90}{Race}}} 
& Concept Algebra~\cite{wang2023concept}
&2.48 & 26.04 \\
& UCE~\cite{gandikota2024unified}
&5.16 & 26.13 \\
& Finetuning Diffusion~\cite{shen2024finetuning}
&4.12 & 25.78 \\
& Interpret Diffusion~\cite{li2024self}
&21.66 & 23.82 \\
& \cellcolor{Gray}EFA (Ours)
&\cellcolor{Gray}\textbf{0.13} & \cellcolor{Gray}\textbf{26.21} \\

\midrule

\parbox[t]{2mm}{\multirow{5}{*}{\rotatebox[origin=c]{90}{G. $\times$ R.}}} 
& Concept Algebra~\cite{wang2023concept}
&2.53 & 26.03 \\
& UCE~\cite{gandikota2024unified}
&6.57 & 25.71 \\
& Finetuning Diffusion~\cite{shen2024finetuning}
&4.26 & 25.71 \\
& Interpret Diffusion~\cite{li2024self}
&41.65 & 22.45 \\
& \cellcolor{Gray}EFA (Ours)
&\cellcolor{Gray}\textbf{0.45} & \cellcolor{Gray}\textbf{26.20} \\

\bottomrule
\end{tabular}
}
\caption{
Evaluation of model preservation using COCO-no-person.
EFA outperforms the baselines in preserving the generation capability of the original model.
}
\vspace{-5mm}
\label{tab:model preservation}
\end{table}

\begin{figure*}[t!]
\centering
\includegraphics[width=\textwidth]{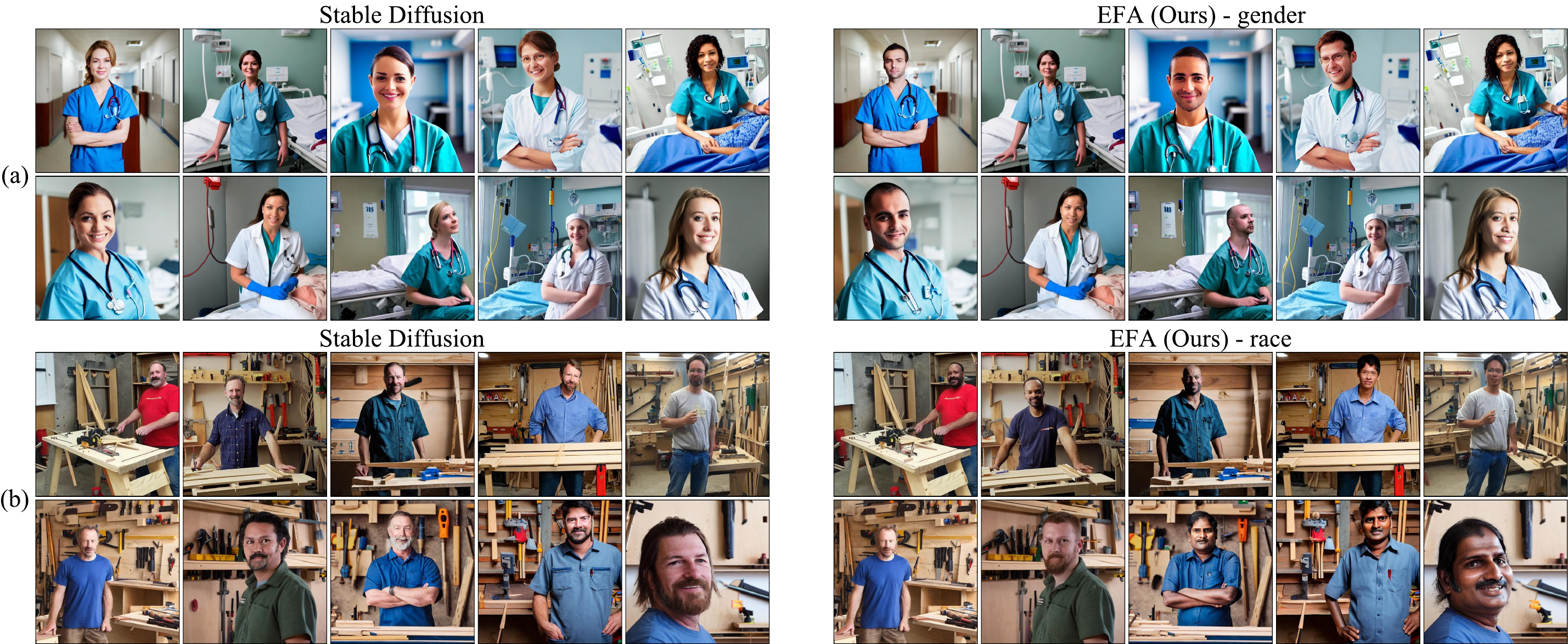}
\caption{
Generated images from the original Stable Diffusion and ours. Ours successfully generates images with a fair distribution of the target bias (\ie, gender and race). (a) and (b) represent the images generated from the prompts describing a nurse in a hospital and a carpenter in a workshop, respectively. The corresponding aligned images were generated using the same random seed. Further qualitative results, including those from the model mitigating gender and race simultaneously, are provided in the Supplementary.
}
\vspace{-3mm}
\label{fig:main_result}
\end{figure*}

\section{Experiments}
\subsection{Experimental settings}
\noindent \textbf{Datasets.}
Following the previous studies~\cite{friedrich2023FairDiffusion, gandikota2024unified, li2024self}, we mainly conducted experiments with gender and racial bias.
Experiments with age bias are provided in Supplementary.
The target attributes set of gender and racial bias is set as $\mathcal{A}_\text{gender}=\{\text{female}, \text{male}\}$ and $\mathcal{A}_\text{race}=\{\text{White}, \text{Black}, \text{Asian}, \text{Indian}\}$, respectively.
For multiple biases, the set of target attributes is defined as $\mathcal{A}_{\text{gender} \times \text{race}}=\mathcal{A}_\text{gender} \times \mathcal{A}_\text{race}$.
We adopt a binary gender perspective to evaluate our method, following prior works~\cite{gandikota2024unified, li2024self}, due to the limitations of existing datasets and evaluation protocols.
Further details of the dataset construction are included in the Supplementary.

\noindent \textbf{Evaluation prompt.}
To assess bias mitigation performance, we utilized 36 occupations from the WinoBias~\cite{zhao2018gender} dataset, which includes stereotypical occupations based on US Department of Labor statistics by following previous studies~\cite{li2024self, gandikota2024unified, orgad2023editing}.
We employed two types of prompt templates for image generation. 
The first set (\ie, $\mathcal{T}_\text{basic}$) follows the prompts used in prior research~\cite{li2024self, gandikota2024unified}, formatted as ``\{style\} of \{occupation\}." 
The style category includes photo, image, picture, headshot, and portrait.
The second set (\ie, $\mathcal{T}_\text{complex}$) includes background components to enable evaluation with more complex non-target attributes. 
Specifically, prompts are structured as ``\{style\} of \{occupation\} in \{background\}," where the background is contextually relevant to the occupation.

To evaluate the overall model preservation performance while ensuring non-target attributes remain unaffected, we assessed whether the model retained its original generation capability when generating images without human presence. 
Specifically, we filtered image-prompt pairs from the COCO-30K~\cite{lin2014microsoft} dataset and constructed a new dataset, COCO-no-person, consisting of 14,926 pairs where no person is present in the corresponding images.

\noindent \textbf{Evaluation metrics.} Our evaluation focuses on (1) bias mitigation, ensuring balanced attribute representation; (2) non-target attribute preservation, maintaining non-target attributes while adjusting target attributes; and (3) model preservation, maintaining the overall generative capability of the original model in non-human contexts.
While model preservation also evaluates the preservation of non-target attributes, we use a distinct term to differentiate it from cases where target attributes should be enhanced.

For evaluating bias mitigation, we use a CLIP classifier to categorize each generated image into attribute-defined groups in $\mathcal{A}_C$ and measure attribute balance using modified deviation ratio~\cite{li2024self}, which is designed to handle an arbitrary number of attributes.
The deviation ratio (DR) is calculated as $\max_{a_i \in \mathcal{A}_C}\frac{|N_{a_i}/N-1/|\mathcal{A}_C||}{1-1/|\mathcal{A}_C|}$, where $N$ and $N_{a_i}$ indicate the total number of generated images and images classified as belonging to class $a_i$, respectively. The lower DR indicates a better balance of attributes. 
We compute scores for each occupation and average them to obtain the final score.

For non-target attribute preservation scores, we generate image pairs using the original Stable Diffusion (SD)~\cite{2022stable} and our approach with the same random seed to evaluate how well non-target attributes are preserved while modifying target attributes. 
First, we segment the non-human regions in the original image and measure how well they are retained in the image generated by ours using Peak Signal-to-Noise Ratio (PSNR) and Learned Perceptual Image Patch Similarity (LPIPS)~\cite{zhang2018unreasonable}. 
To assess overall image preservation, including contextual elements within human regions (\eg, a doctor's stethoscope or white coat), we compute the DINO score, employing DINO's ability to capture low-level and semantically meaningful visual details. The score measures the cosine similarity between the DINO~\cite{Caron_2021_ICCV} embeddings of images generated by SD and our method.

Lastly, for assessing model preservation, we use COCO-no-person dataset.
To measure text fidelity, we use the CLIP-T score, which computes the cosine similarity between CLIP embeddings of the input prompt and the generated images.
Additionally, Fréchet Inception Distance (FID)~\cite{heusel2017gans} is used to evaluate image quality and diversity. 
The FID scores are computed against the generation outputs of the original SD.
Unlike non-target attribute preservation scores, this evaluation ensures that our method does not degrade the generation capability of the pretrained model when bias mitigation is not required.
Note that bold text in all tables indicates the best score.

\noindent \textbf{Implementation details.}
The experiments were conducted based on SD v1.5 followed by previous works~\cite{friedrich2023FairDiffusion, Parihar2024balancingAct, shen2024finetuning, zhou2024association}. 
For evaluation with WinoBias, we generated 160 images per occupation, using random seeds where the original SD has a detectable face.

\begin{figure}[t!]
\centering
\includegraphics[width=\columnwidth]{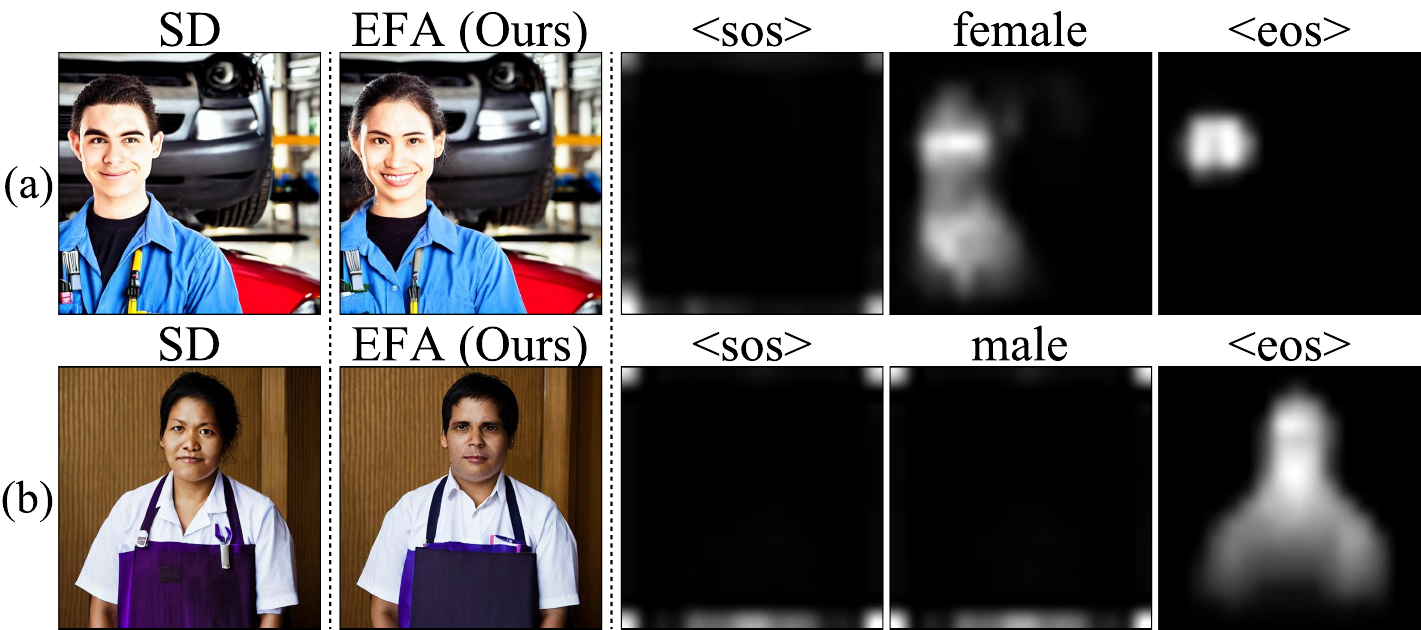}
\caption{
Attention map visualization when mitigating gender bias using EFA. (a) and (b) show the results of applying $\text{EFA}_\text{gender}^\text{female}$ and $\text{EFA}_\text{gender}^\text{male}$, respectively. 
$<$sos$>$ and $<$eos$>$ denote special tokens that mark the beginning and end of a text sequence, respectively.
}
\label{fig:attention map}
\vspace{-4mm}
\end{figure}

\subsection{Experimental results}
\noindent\textbf{Bias mitigation and non-target attribute preservation.}
Table~\ref{tab:recent_comparison} presents the comparative experimental results of our method and baseline models using the WinoBias dataset. 
Overall, models struggle to maintain non-target attributes more in $\mathcal{T}_\text{complex}$ than $\mathcal{T}_\text{basic}$. 
Concept Algebra exhibits low performance in bias mitigation.
Both UCE~\cite{gandikota2024unified} and Finetuning Diffusion~\cite{shen2024finetuning} modify model parameters directly, making it challenging to preserve the non-target attributes consistent with the original SD.
Interpret Diffusion~\cite{li2024self} demonstrates better bias mitigation performance compared to other baselines.
However, it exhibits low non-target attribute preservation scores due to its uniform application of fixed semantic vectors without spatial awareness. 
In contrast, our input-aware architecture adaptively enhances target attributes based on spatial context, enabling effective preservation of non-target attributes.

Our proposed method outperforms existing baselines by achieving a lower DR while having substantially higher non-target attribute preservation scores.
The high preservation scores validate the effectiveness of our approach, showing that the model successfully learns to preserve regions corresponding to non-target attributes.

We hypothesize that the effectiveness in bias mitigation stems from the combination of two key factors: (1) the rich semantic information encoded in text embeddings, which have been pretrained on large-scale datasets, and (2) the capability of U-Net to faithfully reflect these embeddings in the generated output.
By utilizing pretrained text embeddings without altering their inherent semantics, our approach ensures that target attributes are seamlessly integrated into the generation process, while U-Net’s capacity to translate these embeddings into visual features enables precise control over attribute distributions.
The extended experiments on Concept Algebra, provided in the supplementary material, corroborate our hypothesis.
Consequently, our method successfully steers the distribution of target attributes while preserving non-target attributes.

Fig.~\ref{fig:main_result} demonstrates that our method effectively mitigates societal bias, generating outputs that represent a diverse range of gender and race. 
Furthermore, by leveraging regularization terms and utilizing text embeddings disentangled from other attributes, our method successfully preserves non-target attributes, such as overall layout, pose, and background, while effectively mitigating bias.

\noindent\textbf{Model preservation.}
The experimental results on model preservation are presented in Table~\ref{tab:model preservation}. 
Our method achieves a significantly lower FID compared to baseline approaches, indicating superior preservation of both the quality and diversity of SD’s outputs. 
Furthermore, our approach yields CLIP-T scores comparable to those of SD, effectively maintaining the text fidelity of the original model by preserving non-target attributes. 
Qualitative results on model preservation with COCO-no-person and additional experiments using COCO-30K are provided in the Supplementary.

\section{Analysis}
\subsection{Effect of regularization and segmentation mask}
\label{subsec:anal_loss}
\definecolor{Gray}{gray}{0.92}

\begin{table}[t]
\centering
\resizebox{1.0\columnwidth}{!}{
\setlength{\tabcolsep}{0.6em}
\def\arraystretch{1.0}%
\begin{tabular}{ l | c | c c c }
\toprule 

\multicolumn{1}{c|}{\multirow{2}{*}{}}

& \multicolumn{1}{c|}{{Bias}}  
& \multicolumn{3}{c}{{Non-target attribute P.}}
\\
\cmidrule(lr){2-2} \cmidrule(lr){3-5} 

& DR $\downarrow$
& PSNR $\uparrow$
& LPIPS $\downarrow$
& DINO $\uparrow$

\\
\midrule

Original SD & 0.71 & - & - \\
EFA w/o $\mathcal{L}_\text{reg}^\text{trg}$ 
& 0.07 & 27.76 & 0.0527 & 0.890 \\

EFA w/o seg. mask 
&  \textbf{0.04} & 29.31 & 0.0562 & 0.888 \\

\rowcolor{Gray}EFA (Ours) 
&  0.06 & \textbf{32.52} & \textbf{0.0411} & \textbf{0.916}  \\

\bottomrule
\end{tabular}
}
\caption{
Effectiveness of the regularization loss and the segmentation mask in bias mitigation and non-target attribute preservation. 
P. and seg. indicate preservation and segmentation, respectively.
}
\vspace{-3mm}
\label{tab:ablation_loss}
\end{table}
As shown in Table~\ref{tab:ablation_loss}, we conduct an ablation study to examine the effects of regularization losses and segmentation masks on non-target attribute preservation. 
For the experiments, we train models to mitigate gender bias, and the evaluation is performed using $\mathcal{T}_\text{basic}$.

When training without segmentation masks, we exclude $\mathcal{L}_\text{reg}^\text{cf}$ in training and apply $\mathcal{L}_\text{recon}$ to the entire image. 
Since there is no constraint limiting target attribute enhancement, we observe a slight improvement in bias mitigation performance. 
However, without segmentation masks to distinguish human and non-human regions during training, EFA struggles to localize them precisely, leading to reduced preservation of non-target attributes at inference.
Nonetheless, EFA still outperforms baselines even without segmentation masks. 
We conjecture that this is because counterfactual attributes are primarily activated in bias-relevant regions due to their semantic similarity in cross-attention, which encourages EFA to enhance the target attribute in those regions while ignoring the counterfactual prompt’s influence. 
This allows EFA to approximate the localization of bias-relevant regions without relying on fine-grained masks of the target attribute.
Also, when $\mathcal{L}_\text{reg}^\text{trg}$ is not applied, the non-target attribute preservation performance deteriorates, even though the bias mitigation performance remains similar.
These results highlight the necessity of the regularization mechanisms to preserve non-target attributes across diverse scenarios where the target attribute appears at different extents in the input.

\subsection{Analyzing the influence of resolution in EFA}
\label{subsec:anal_layer}
\definecolor{Gray}{gray}{0.92}

\begin{table}[t]
\centering
\resizebox{1.0\columnwidth}{!}{
\setlength{\tabcolsep}{0.65em}
\def\arraystretch{1.0}%
\begin{tabular}{ l | c | c c c }
\toprule 

\multicolumn{1}{c|}{\multirow{2}{*}{}}

& \multicolumn{1}{c|}{{Bias}}  
& \multicolumn{3}{c}{{Non-target attribute P.}}
\\
\cmidrule(lr){2-2} \cmidrule(lr){3-5} 
resolution of input
& DR $\downarrow$
& PSNR $\uparrow$
& LPIPS $\downarrow$
& DINO $\uparrow$

\\
\midrule

\rowcolor{Gray}16$\times$16 (Ours) &  \textbf{0.06} & \textbf{32.52} & 0.0411 & \textbf{0.916}\\
16$\times$16 - 32$\times$32 &  \textbf{0.06} & 31.79 & \textbf{0.0407} & 0.915\\
16$\times$16 - 64$\times$64 &  \textbf{0.06} & 30.65 & 0.0438 & 0.911\\
\bottomrule
\end{tabular}
}
\caption{
Impact of input resolution employed by EFA. 
Extending EFA to higher-resolution layers compromises the preservation of non-target attributes.
P. denotes preservation.
}
\vspace{-5mm}
\label{tab:ablation_layer}
\end{table}
To investigate the impact of input resolution, we integrated EFA into additional layers within the up-blocks while training a model for gender bias mitigation.
The performance on $\mathcal{T}_\text{basic}$ is reported in Table~\ref{tab:ablation_layer}, where the row highlighted in gray represents our proposed method.
When EFA is applied to higher-resolution features, the deviation ratio remains nearly unchanged, while PSNR and DINO scores exhibit a slight decline. 
This suggests that lower-resolution layers contain richer semantic information, which is crucial for attribute modification.
Moreover, extending EFA to multiple layers inevitably introduces additional noise, potentially compromising the preservation of non-target attributes.

\subsection{Attention map visualization}
\label{subsec:attention_vis}
In Fig.~\ref{fig:attention map}, we visualize the attention maps predicted by EFA when the target attribute is enhanced.
EFA assigns high values to the human region, effectively enabling gender transformation while preserving the generation capability of the original model, as the impact on accessories, clothing style, and background details remains minimal.

\section{Conclusion}
In this work, we addressed attribute entanglement in diffusion-based T2I models, where bias mitigation often distorts non-target attributes, leading to unintended distribution shifts.
We propose EFA, which mitigates human-centric biases while preserving non-target attributes.
During training, EFA is guided to attend regions with high semantic relevance to counterfactual prompts, constrained by human segmentation masks.
At inference, EFA randomly samples a target attribute and alters the cross-attention in selected layers to ensure a fair distribution of target attributes in the outputs.
Experimental results verify that EFA outperforms prior methods in mitigating bias while preserving the original model’s output distribution and generative capacity.
\clearpage

\section*{Acknowledgments}
This work was supported by Institute for Information \& communications Technology Planning \& Evaluation(IITP) grant funded by the Korea government(MSIT) (RS-2019-II190075, Artificial Intelligence Graduate School Program(KAIST)), Institute of Information \& communications Technology Planning \& Evaluation (IITP) grant funded by the Korea government (MSIT) (No.RS-2021-II212068, Artificial Intelligence Innovation Hub), and the National Research Foundation of Korea(NRF) grant funded by the Korea government(MSIT)(No. 2022R1A5A7083908).
{
    \small
    \bibliographystyle{ieeenat_fullname}
    \bibliography{main}
}

\clearpage
\setcounter{page}{1}
\maketitlesupplementary

\section{Implementation details}
\label{sec:supp_implement_details}
\noindent\textbf{Architecture.}
The spatial information is crucial for distinguishing between target and non-target attributes.
Therefore, we adopt convolutional networks as our backbone, given their ability to capture and process spatial structural information.
Specifically, AP consists of three 2D convolutional layers and two SiLU activation functions. 
AP predicts attention values for token embeddings of target attributes for each attention head, and the token embeddings of the target attribute include special tokens (\ie, start-of-sequence and end-of-sequence tokens). 
The target attribute and user prompt are separately processed during the text embedding stage.
To leverage rich semantic information, EFA is applied to selected up-block layers where the input resolution is low. Specifically, the input to EFA is given by $\textbf{z}_t\in\mathbb{R}^{B\times S\times D}$, where $B$, $S$, and $D$ denote the batch size, spatial resolution ($S=16\times16$), and feature dimension, respectively.
Note that applying EFA does not change the input or output shapes of the cross-attention module, since the target attribute features are added only to the key and value matrices while leaving the overall architecture intact.

\noindent\textbf{Training and Inference.}
For experiments, NVIDIA RTX 3090, A6000, or V100 GPUs are employed.
The models for gender, race, and their intersectional biases were trained for 20 epochs, while the model incorporating age as an attribute was optimized for 10 epochs.
Hyperparameters were selected based on validation set performance. 
When training with multiple target attributes, we independently tuned the hyperparameters for each attribute to ensure optimal learning.
During inference, we set the timestep for image generation to 50 for all baselines.

\noindent\textbf{Evaluation.}
To measure the deviation ratio, we utilized a CLIP classifier to predict gender and race from the images.
Following previous work~\cite{li2024self}, `a woman' and `a man' were used for gender prediction, and template `a \{race\}-race person' was employed for race prediction.
For quantitative comparisons, we adhered to the optimal settings established by baseline models and followed the implementation details provided in the publicly available code.

To evaluate non-target attribute preservation, we use paired images generated by the original Stable Diffusion (SD) and the bias mitigation method under identical conditions.
When calculating PSNR and LPIPS, if the Grounded SAM2 failed to detect a person and thus could not generate a segmentation mask, the corresponding image was excluded from the evaluation. 
The excluded images accounted for less than 1\% of the total evaluation set.
For qualitative results, we ensured randomness across all baselines by aligning the scheduler and fine-grained settings with the default SD configuration.

\section{Dataset construction details}
\label{sec:supp_dataset_details}
This section elaborates on the details of the datasets utilized during the training, validation, and testing phases.

\subsection{Training dataset construction}
For training datasets, we generated images using SD v1.5 with attribute-specific prompts, detailed in Table~\ref{tab:supp_data_construction}. 
The guidance scale was set to 7.5, and inference steps to 50.
For single-bias models, 2K images were generated per attribute.
Multi-bias models defined their attribute sets as the Cartesian product of single-bias attribute sets, generating 1K images for each combination of attributes.
After obtaining the images, human segmentation masks were extracted using the Grounded SAM2 model. Images without detected persons ($<$ 1\%) were excluded from training.
\definecolor{Gray}{gray}{0.92}

\begin{table*}[t]
\centering
\resizebox{0.8\textwidth}{!}{
\setlength{\tabcolsep}{1.0em}
\def\arraystretch{1.0}%
\begin{tabular}{ l | c c }
\toprule

Target Bias & Set of Target Attributes & Prompt Template \\

\midrule
gender & $\mathcal{A}_\text{gender}=\{\text{female}, \text{male}\}$ & a \{gender\} person \\
race & $\mathcal{A}_\text{race}=\{\text{White}, \text{Black}, \text{Asian}, \text{Indian}\}$ & a/an \{race\}-race person \\
gender, race & $\mathcal{A}_\text{gender} \times \mathcal{A}_\text{race}$ & a/an \{race\}-race \{gender\} person \\
gender, race, age & $\mathcal{A}_\text{gender} \times \mathcal{A}_\text{race} \times \{\text{young}, \text{old}\}$ & a/an \{age\} \{race\}-race \{gender\} person \\

\bottomrule
\end{tabular}
}
\caption{
Set of target attributes for each target bias and prompt templates used for image generation.
}
\label{tab:supp_data_construction}
\end{table*}

\subsection{Assignment of contextual backgrounds to professions}
To facilitate evaluation in more complex backgrounds, we extended the original prompt templates by incorporating background information.
To systematically assign appropriate backgrounds to each profession, we employed a large language model (LLM). 
Specifically, we utilized GPT-4o mini, providing it with the profession list from WinoBias alongside the following query:
\textit{``Can you write a place for each profession in the following list? Please provide the answers in the format `\{profession\}:\{place\}' to match each profession with a place. I will use the answers to create SD prompts in the format `\{profession\} in \{place\}'"}.

This approach enabled us to systematically extract contextually relevant locations for each profession. The resulting profession-specific background assignments are as follows:
\{Attendant: train station,
Cashier: supermarket,
Teacher: classroom,
Nurse: hospital,
Assistant: office,
Secretary: corporate office,
Cleaner: office building,
Receptionist: hotel lobby,
Clerk: retail store,
Counselor: counseling office,
Designer: studio,
Hairdresser: salon,
Writer: library,
Housekeeper: hotel,
Baker: bakery,
Librarian: library,
Tailor: tailoring shop,
Driver: bus,
Supervisor: construction site,
Janitor: school,
Cook: restaurant,
Laborer: factory,
Construction worker: construction site,
Developer: tech company office,
Carpenter: workshop,
Manager: office building,
Lawyer: law firm,
Farmer: farm,
Salesperson: mall,
Physician: clinic,
Guard: security post,
Analyst: corporate office,
Mechanic: auto repair shop,
Sheriff: sheriff's office,
CEO: corporate headquarters,
Doctor: hospital\}

\subsection{Validation dataset}
To identify the optimal hyperparameters during the training process, we employed $\mathcal{T}_\text{basic}$ using 20 professions that do not overlap with those in WinoBias. The validation professions include
\textit{Maid,
Therapist,
Author,
Model,
Caregiver,
Florist,
Laundry worker,
Telemarketer,
Wedding planner,
Yoga instructor,
Bioengineer,
Plumber,  
Athlete,
Bartender,
Industrialist,
Judge, 
Woodworker,
Security,
Pilot,
Firefighter}.

\section{Extending EFA to handle multiple biases}
The real-world biases are often multifaceted, involving multiple attributes such as gender and race.
To address this, we extend EFA to simultaneously handle multiple bias concepts by defining the attribute space as the Cartesian product of individual attribute sets.
Specifically, when considering both $C_1$ (\eg, gender) and $C_2$ (\eg, race) biases, we define the expanded attribute space as $\mathcal{A}_{C_1\times C_2} = \mathcal{A}_{C_1} \times \mathcal{A}_{C_2}$.
This formulation allows EFA to learn attribute-specific modifications that account for both $C_1$ and $C_2$ simultaneously.

To achieve this, EFA predicts attention values corresponding to multiple attributes associated with $C_1$ and $C_2$ (\eg, female, indian).
This process follows the same principle as the single-bias EFA, ensuring that both attributes are faithfully reflected in the generated output without disrupting non-target attributes.

The overall training framework remains identical to the single-bias case, with the primary distinction being that EFA now predicts and incorporates multiple attribute-specific embeddings.
By handling multiple biases simultaneously, our approach ensures that the generated outputs exhibit balanced attribute representation across multiple dimensions of fairness.

\section{Simultaneous mitigation of gender, racial, and age biases}
This section presents experimental results with multiple biases that include gender, race, and age.
We trained our model to address these biases simultaneously and demonstrate its effectiveness.
In evaluation, to predict age from an image, we utilized the age prediction model employed in the previous work~\cite{shen2024finetuning}.
Table~\ref{tab:supp_age} presents the performance of our model on the WinoBias dataset. On the COCO-no-person dataset, our method achieved an FID of 0.53 and a CLIP-T score of 26.07.

These results demonstrate that our approach successfully mitigates the multiple biases of the pretrained model while minimizing distortions in non-target attributes.
Furthermore, even compared to baseline methods trained to address a single bias, our method effectively tackles multiple biases while achieving better performance in both non-target attribute preservation and model preservation, with minimal compromise in text fidelity.

\definecolor{Gray}{gray}{0.92}

\begin{table}[t]
\centering
\resizebox{\columnwidth}{!}{
\setlength{\tabcolsep}{0.8em}
\def\arraystretch{1.0}%
\begin{tabular}{ l l | c c c c }
\toprule

&
& \multicolumn{1}{c}{{Bias}}  
& \multicolumn{3}{c}{{Non-target attribute P.}}
\\
\cmidrule(lr){3-3} \cmidrule(lr){4-6}
& Method
& DR $\downarrow$
& PSNR $\uparrow$
& LPIPS $\downarrow$
& DINO $\uparrow$

\\
\midrule
\parbox[t]{2mm}{\multirow{2}{*}{\rotatebox[origin=c]{90}{$\mathcal{T}_\text{b}$}}} &
Original SD 
& 0.45 & - & - & - \\

& \cellcolor{Gray}EFA (Ours) 
& \cellcolor{Gray}\textbf{0.05} & \cellcolor{Gray}\textbf{24.69} & \cellcolor{Gray}\textbf{0.0740} & \cellcolor{Gray}\textbf{0.838} \\

\midrule

\parbox[t]{2mm}{\multirow{2}{*}{\rotatebox[origin=c]{90}{$\mathcal{T}_\text{c}$}}} &

Original SD 
& 0.46 & - & - & - \\

& \cellcolor{Gray}EFA (Ours)
& \cellcolor{Gray}\textbf{0.08} & \cellcolor{Gray}\textbf{23.51} & \cellcolor{Gray}\textbf{0.0811} & \cellcolor{Gray}\textbf{0.900}\\

\bottomrule
\end{tabular}
}
\caption{
P. is the abbreviation for preservation. 
$\mathcal{T}_\text{b}$ and $\mathcal{T}_\text{c}$ indicates $\mathcal{T}_\text{basic}$ and $\mathcal{T}_\text{complex}$, respectively.
}
\label{tab:supp_age}
\end{table}
\definecolor{Gray}{gray}{0.92}

\begin{table*}[t]
\centering
\resizebox{\textwidth}{!}{
\setlength{\tabcolsep}{1.0em}
\def\arraystretch{1.0}%
\begin{tabular}{ l l | c c c c | c  c c c }
\toprule 

\multicolumn{2}{c|}{\multirow{2}{*}{}}

&\multicolumn{4}{c|}{{$\mathcal{T}_\text{basic}$}}  
&\multicolumn{4}{c}{{$\mathcal{T}_\text{complex}$}}  
\\
\cmidrule(lr){3-6} \cmidrule(lr){7-10}
&
& \multicolumn{1}{c}{{Bias}}  
& \multicolumn{3}{c|}{{Non-target attribute P.}}
& \multicolumn{1}{c}{{Bias}}  
& \multicolumn{3}{c}{{Non-target attribute P.}}
\\
\cmidrule(lr){3-3} \cmidrule(lr){4-6}
\cmidrule(lr){7-7} \cmidrule(lr){8-10}
& Method
& DR $\downarrow$
& PSNR $\uparrow$
& LPIPS $\downarrow$
& DINO $\uparrow$
& DR $\downarrow$
& PSNR $\uparrow$
& LPIPS $\downarrow$
& DINO $\uparrow$

\\
\midrule
\parbox[t]{2mm}{\multirow{4}{*}{\rotatebox[origin=c]{90}{Gender}}} &
Original SD 
& 0.71 & - & - & - 
& 0.71 & - & - & - \\

& Concept Algebra~\cite{wang2023concept} 
& 0.59 & 21.10 & 0.1169 & 0.834
& 0.69 & 16.69 & 0.1852 & 0.839 \\

& Concept Algebra+~\cite{wang2023concept} 
& \textbf{0.02} & 19.21& 0.1579 & 0.734
& \textbf{0.02} & 15.05& 0.2397 & 0.752 \\

& \cellcolor{Gray}EFA (Ours) 
& \cellcolor{Gray}0.06 & \cellcolor{Gray}\textbf{32.52} & \cellcolor{Gray}\textbf{0.0411} & \cellcolor{Gray}\textbf{0.916}
& \cellcolor{Gray}0.06 & \cellcolor{Gray}\textbf{29.70} & \cellcolor{Gray}\textbf{0.0492} & \cellcolor{Gray}\textbf{0.941} \\

\midrule

\parbox[t]{2mm}{\multirow{4}{*}{\rotatebox[origin=c]{90}{Race}}} &

Original SD 
& 0.60 & - & - & - 
& 0.55 & - & - & - \\

& Concept Algebra~\cite{wang2023concept}
&0.64 & 21.47 & 0.1164 & 0.839
&0.58 & 16.62 & 0.1888 & 0.841 \\

& Concept Algebra+~\cite{wang2023concept}
& 0.10 & 18.09 & 0.1817 & 0.708
&0.12 & 14.22 &	0.2653 & 0.723 \\

& \cellcolor{Gray}EFA (Ours)
& \cellcolor{Gray}\textbf{0.04} & \cellcolor{Gray}\textbf{30.93} & \cellcolor{Gray}\textbf{0.0353} & \cellcolor{Gray}\textbf{0.938}
& \cellcolor{Gray}\textbf{0.06} & \cellcolor{Gray}\textbf{28.55} & \cellcolor{Gray}\textbf{0.0421} & \cellcolor{Gray}\textbf{0.958}\\

\midrule

\parbox[t]{2mm}{\multirow{4}{*}{\rotatebox[origin=c]{90}{G. $\times$ R.}}} &
Original SD 
& 0.56 & - & - & - 
& 0.50 & - & - & - \\
& Concept Algebra~\cite{wang2023concept}
& 0.51 & 20.12 & 0.1325 & 0.805
& 0.47 & 15.88 & 0.2031 & 0.818 \\

& Concept Algebra+~\cite{wang2023concept}
& 0.09 & 16.70 & 0.2108 & 0.645
& 0.09 & 13.43 & 0.2930 & 0.674 \\

& \cellcolor{Gray}EFA (Ours)
& \cellcolor{Gray}\textbf{0.03} & \cellcolor{Gray}\textbf{25.58} & \cellcolor{Gray}\textbf{0.0684} & \cellcolor{Gray}\textbf{0.853}
& \cellcolor{Gray}\textbf{0.05} & \cellcolor{Gray}\textbf{23.78} & \cellcolor{Gray}\textbf{0.0795} & \cellcolor{Gray}\textbf{0.903} \\

\bottomrule
\end{tabular}
}
\caption{
Quantitative comparison of Concept Algebra, Concept Algebra+, and EFA (Ours). P. is the abbreviation for preservation. 
While the bias mitigation performance of Concept Algebra+ is comparable to that of EFA, EFA demonstrates significantly superior non-target attribute preservation compared to the baselines.
}
\label{tab:supp_CAplus}
\end{table*}
\definecolor{Gray}{gray}{0.92}

\begin{table}[t]
\centering
\resizebox{\columnwidth}{!}{
\setlength{\tabcolsep}{1.0em}
\def\arraystretch{1.0}%
\begin{tabular}{ l l | c  c }
\toprule

& Method
& FID $\downarrow$
& CLIP-T $\uparrow$
\\
\midrule
-
&Original SD
&- & 26.17 \\
\midrule

\parbox[t]{2mm}{\multirow{3}{*}{\rotatebox[origin=c]{90}{Gender}}}
& Concept Algebra~\cite{wang2023concept}
&2.41 & 26.02 \\
& Concept Algebra+~\cite{wang2023concept}
& 3.64 & 25.75 \\
& \cellcolor{Gray}EFA (Ours) 
& \cellcolor{Gray}\textbf{0.23} & \cellcolor{Gray}\textbf{26.03} \\

\midrule

\parbox[t]{2mm}{\multirow{3}{*}{\rotatebox[origin=c]{90}{Race}}} 
& Concept Algebra~\cite{wang2023concept}
&2.48 & 26.04 \\
& Concept Algebra+~\cite{wang2023concept}
& 7.12 & 25.46 \\
& \cellcolor{Gray}EFA (Ours)
&\cellcolor{Gray}\textbf{0.13} & \cellcolor{Gray}\textbf{26.21} \\

\midrule

\parbox[t]{2mm}{\multirow{3}{*}{\rotatebox[origin=c]{90}{G. $\times$ R.}}} 
& Concept Algebra~\cite{wang2023concept}
&2.53 & 26.03 \\
& Concept Algebra+~\cite{wang2023concept}
&10.33 & 25.03 \\
& \cellcolor{Gray}EFA (Ours)
&\cellcolor{Gray}\textbf{0.45} & \cellcolor{Gray}\textbf{26.20} \\

\bottomrule
\end{tabular}
}
\caption{
Evaluation of model preservation in terms of image quality and text fidelity using COCO-no-person.
Concept Algebra+ shows lower model preservation and greater deviation from the original output compared to standard Concept Algebra and EFA.
}
\label{tab:supp_CAplus_coco}
\end{table}

\section{Efficiency Analysis}
We evaluated EFA that simultaneously mitigates gender, race, and age biases on an NVIDIA A100 GPU. 
The number of parameters increased by 9.4\% compared to the original SD (1066M $\to$ 1166M), and the additional inference time is minimal at 0.29 seconds (3.69s $\to$ 3.98s).
This overhead could potentially be reduced through implementation optimizations and exploration of lightweight variants of EFA for deployment in resource-constrained environments.

\section{Extending Concept Algebra with target attribute guidance}

Concept Algebra generates the output from the original prompt (\eg ``a portrait of a mathematician") with the target concept (\eg Fauvism style) by manipulating its representation in the subspace of the concept spans (\eg a subspace of style).
In this way, Concept Algebra can modify only the attributes related to the target concept, maintaining the others. 
In the original paper, they utilize ``person'' as the target concept to obtain the desired gender distribution in its identified subspace, mitigating gender bias.

However, since our method assumes the presence of predefined target attributes (\eg female and male), we can utilize these target attributes, instead of ``person'', as the target concept for Concept Algebra.
This can provide more explicit guidance to manipulate the outputs corresponding to the target attributes.
We named this modified version as Concept Algebra+.
For example, we mitigate gender bias using ``female person" and ``male person" to guide the gender distribution of the original prompt.

Table~\ref{tab:supp_CAplus} and Table~\ref{tab:supp_CAplus_coco} present the results of Concept Algebra+ on the WinoBias and COCO-no-person datasets, respectively.
Concept Algebra+ demonstrates significantly improved bias mitigation performance compared to the original Concept Algebra.
This result suggests that directly guiding the desired distribution with specific target attributes, combined with the semantic information encoded in their text embeddings and U-Net, enhances bias mitigation by providing explicit and accurate instructions for adjusting bias.

However, providing explicit guidance based on target attributes introduces greater deviations in the model’s output, leading to a decline in non-target attribute preservation scores compared to the standard Concept Algebra.
In the COCO-no-person dataset, Concept Algebra+ exhibits an output distribution that deviates further from the original SD output compared to standard Concept Algebra, which is also reflected in the decrease in CLIP-T scores.

In contrast, EFA outperforms Concept Algebra+ in non-target attribute preservation while maintaining superior bias mitigation capabilities.
Furthermore, on the COCO-no-person dataset, our method effectively preserves the original SD output while maintaining CLIP-T scores.
These results indicate that our approach effectively provides precise guidance for enhancing target attributes while minimizing interference with non-target attributes.

\begin{figure*}[t!]
\centering
\includegraphics[width=\textwidth]{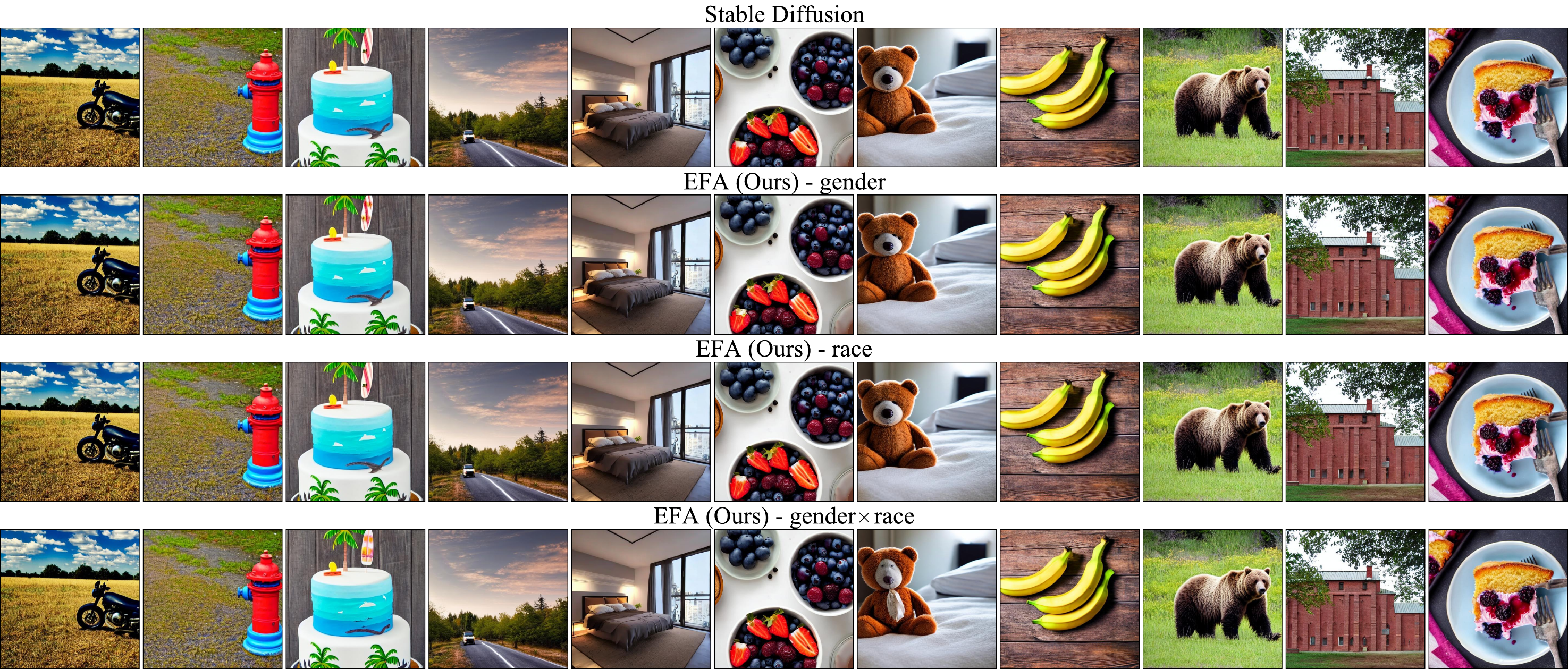}
\caption{
Images generated by SD and our method using prompts from the COCO-no-person dataset. Corresponding aligned images were generated with the same random seed.
EFA maintains the original generation capacity of SD by preserving non-target attributes.
}
\label{fig:supp-coco-no-person}
\end{figure*}

\section{Qualitative Evaluation of Bias Mitigation Approaches}
\begin{figure*}[t!]
\centering
\includegraphics[width=0.8\textwidth]{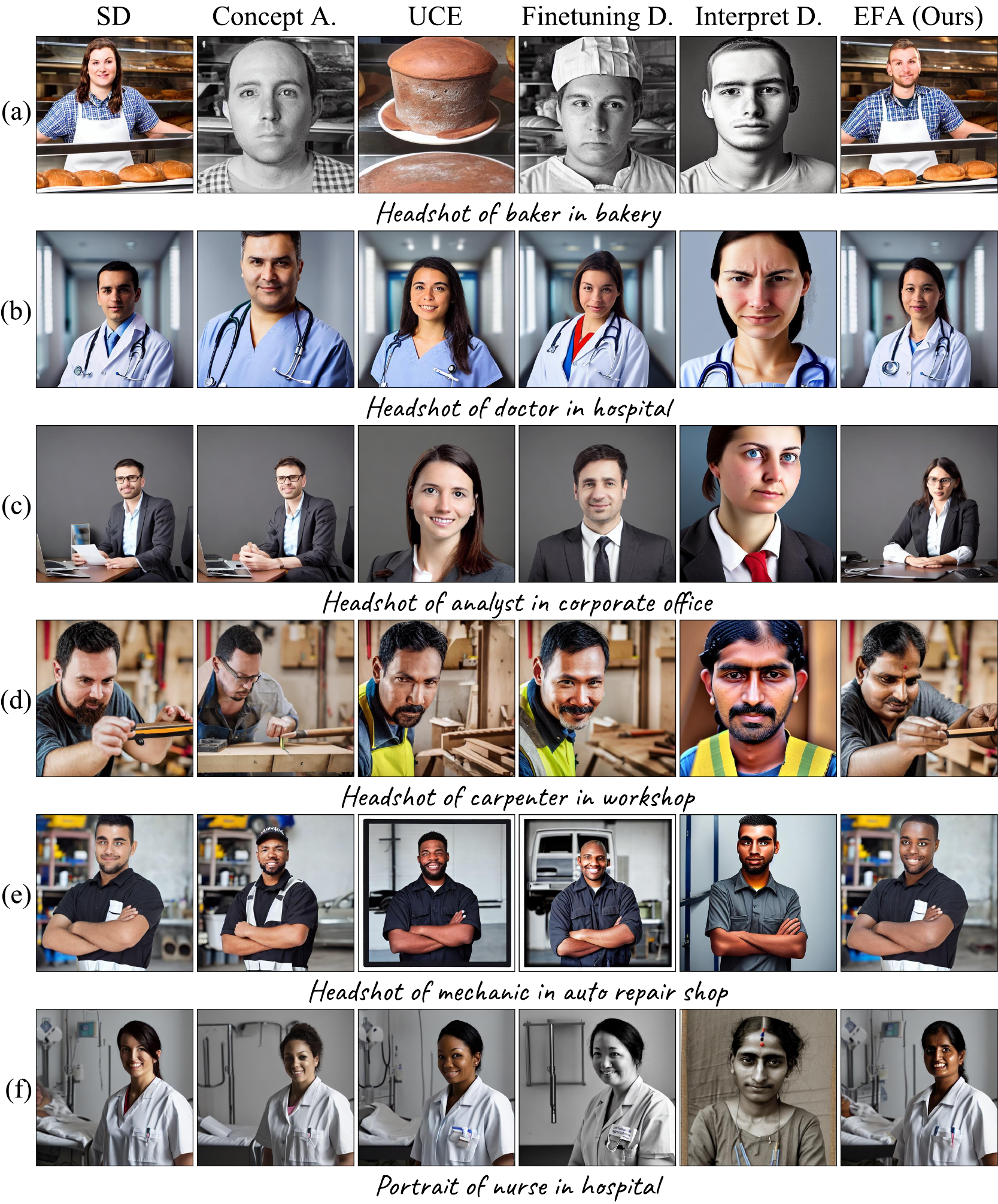}
\caption{
Images generated by SD, baseline methods, and our approach using prompts from the WinoBias dataset. A. and D. are the abbreviations of Algebra and Diffusion, respectively. (a), (b), and (c) represent the results of models addressing gender bias, while (d), (e), and (f) show the results of models addressing racial bias. The corresponding aligned images were generated using the same random seed.
EFA generates diverse genders and races while better preserving non-target attributes compared to other methods.
}
\label{fig:supp-qual-compare}
\end{figure*}

Fig.~\ref{fig:supp-qual-compare} presents qualitative results of SD, baseline methods, and our approach.
(a), (b), and (c) show the results of models addressing gender bias, while (d), (e), and (f) show the results of models addressing racial bias. 
The images in corresponding positions were generated using the same random seed.
Previous methods often fail to maintain the original generation quality of the model, as they alter the layout or fail to preserve background details.
In particular, while Interpret Diffusion demonstrates strong performance in bias mitigation, it often loses visual details that convey occupational characteristics, as seen in examples such as (f).
In contrast, our method produces images with diverse genders and races while successfully maintaining layout, background details, and occupation-relevant visual elements such as a stethoscope and doctor's coat.

\definecolor{Gray}{gray}{0.92}

\begin{table}[t]
\centering
\resizebox{\columnwidth}{!}{
\setlength{\tabcolsep}{1.0em}
\def\arraystretch{1.0}%
\begin{tabular}{ l l | c  c }
\toprule

& Method
& FID $\downarrow$
& CLIP-T $\uparrow$
\\
\midrule
-
&Original SD
&- & 26.31 \\
\midrule

\parbox[t]{2mm}{\multirow{5}{*}{\rotatebox[origin=c]{90}{Gender}}}
& Concept Algebra~\cite{wang2023concept}
&1.32 & 26.08 \\

& UCE~\cite{gandikota2024unified}
&7.73 & 25.14 \\

& Finetuning Diffusion~\cite{shen2024finetuning}
&1.09 & 25.90 \\

& Interpret Diffusion~\cite{li2024self}
&14.32 & 24.76 \\

& \cellcolor{Gray}EFA (Ours) 
& \cellcolor{Gray}0.32 & \cellcolor{Gray}26.19 \\

\midrule

\parbox[t]{2mm}{\multirow{5}{*}{\rotatebox[origin=c]{90}{Race}}} 
& Concept Algebra~\cite{wang2023concept}
&1.36 & 26.17 \\

& UCE~\cite{gandikota2024unified}
&3.61 & 26.15 \\

& Finetuning Diffusion~\cite{shen2024finetuning}
&2.75 & 25.83 \\

& Interpret Diffusion~\cite{li2024self}
&20.71 & 23.71 \\

& \cellcolor{Gray}EFA (Ours)
&\cellcolor{Gray}0.25 & \cellcolor{Gray}26.28
\\

\midrule

\parbox[t]{2mm}{\multirow{5}{*}{\rotatebox[origin=c]{90}{G. $\times$ R.}}} 
& Concept Algebra~\cite{wang2023concept}
& 1.39 & 26.12 \\

& UCE~\cite{gandikota2024unified}
& 4.73 & 25.69 \\

& Finetuning Diffusion~\cite{shen2024finetuning}
& 2.85 & 25.72 \\

& Interpret Diffusion~\cite{li2024self}
& 39.04 & 22.01 \\

& \cellcolor{Gray}EFA (Ours)
&\cellcolor{Gray}0.57 &\cellcolor{Gray}26.00 \\

\bottomrule
\end{tabular}
}
\caption{
Evaluation of model preservation in terms of image quality and text fidelity using COCO-30K.
Compared to baselines, EFA better retains the original model’s generation quality.
}
\label{tab:supp_coco}
\end{table}

\section{Additional quantitative and qualitative results}

\begin{figure*}[t!]
\centering
\includegraphics[width=\textwidth]{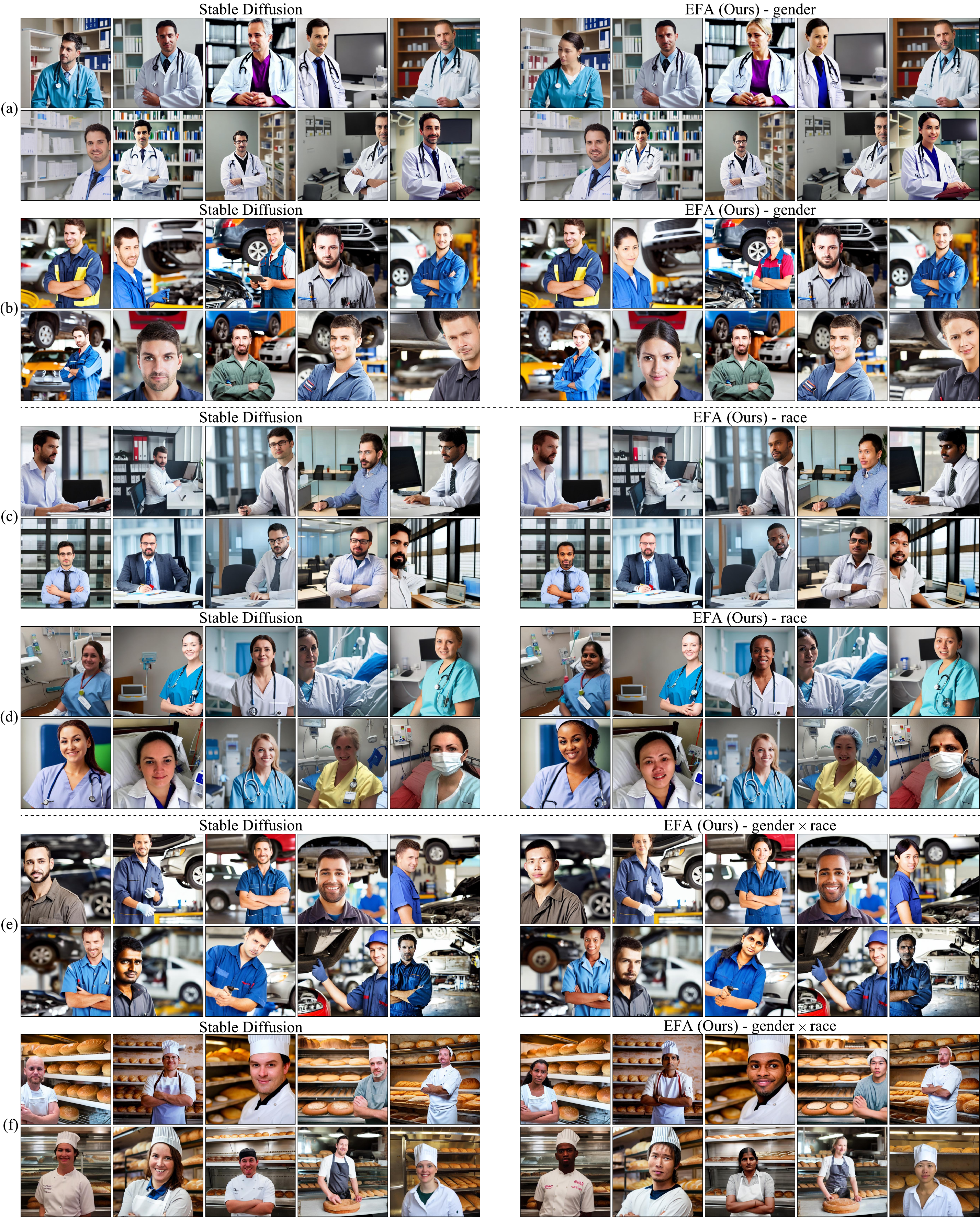}
\caption{
Generated images from original SD and ours. Our approaches successfully generate images with a fair distribution of the target bias. 
(a), (b), (c), (d), (e), and (f) represent the images of a physician, mechanic, analyst, nurse, mechanic, and baker, respectively. The images in the corresponding positions were generated using the same random seed.
}
\label{fig:supp_additional_qual}
\end{figure*}

\noindent\textbf{WinoBias.}
Table~\ref{tab:supp_deviation_ratio} presents the deviation ratio per occupation for SD and our method, while Fig.~\ref{fig:supp_additional_qual} provides additional generated images from our models designed to mitigate biases related to gender, race, and their intersection (gender $\times$ race), comparing them with those from SD.

\begin{figure}[t]
\centering
\includegraphics[width=\linewidth]{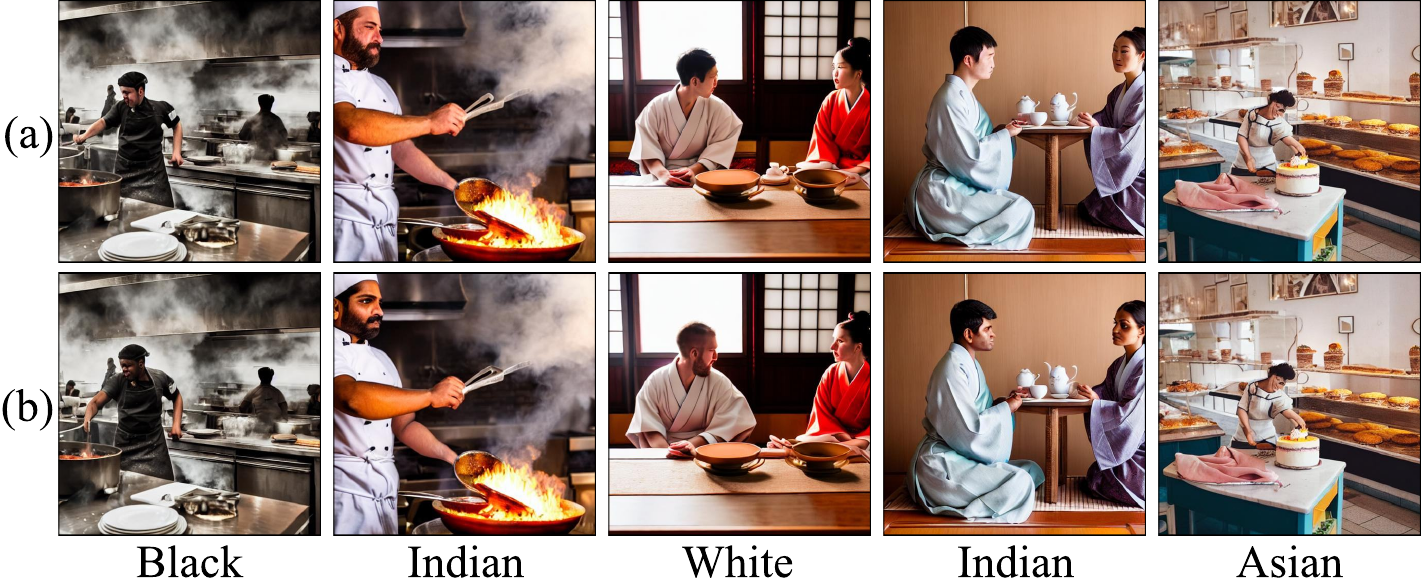}
\caption{
Images generated by SD and our method in complex scenes. (a) and (b) show the results of SD and our method that mitigates racial bias, respectively. The last row indicates the target attributes, and the corresponding aligned images were generated using the same random seed. Our method (EFA) successfully preserves non-target attributes while generating complex scenes.
}
\label{fig:supp-complexscene}
\end{figure}

\noindent\textbf{COCO-30K.}
Table~\ref{tab:supp_coco} presents results for the entire COCO-30K dataset. 
Specifically, we measure the CLIP-T and FID scores of both the baselines and our method. 
The results on COCO-30K exhibit a similar trend to those on COCO-no-person. 
As shown in Fig.~\ref{fig:supp-coco-no-person}, our method effectively mitigates bias while minimizing changes in non-target attributes. 
As a result, it achieves CLIP-T scores comparable to those of the original SD.

\noindent\textbf{Complex Scenes.}
Fig.~\ref{fig:supp-complexscene} presents results on complex scenes containing multiple individuals, diverse objects, and cultural contexts.
Our EFA model for mitigating race bias successfully introduces diverse racial attributes while preserving non-target attributes and maintaining complex scene structure.
However, in cases where the SD fails to generate clear facial features (\eg, the last column of Fig.~\ref{fig:supp-complexscene}), our method struggles to accurately localize and interpret the facial region, thereby limiting its ability to address race-related cues.

\section{Limitations and Future Work}
Assuming a fixed type of bias may limit the general applicability of debiasing approaches in real-world scenarios. Recent advances in automatic bias identification for generative models~\cite{chinchure2025tibet} offer orthogonal solutions that could complement our work.
Integrating such techniques into our framework presents a promising direction for future research.
Furthermore, while our work primarily focuses on human-centric biases, a more systematic investigation into non-human-centric biases remains a valuable topic for further exploration.

\definecolor{Gray}{gray}{0.92}

\begin{table*}[t]
\centering
\resizebox{\textwidth}{!}{
\setlength{\tabcolsep}{0.7em}
\def\arraystretch{1.1}%
\begin{tabular}{ r | c c | c c | c c | c c | c c | c c | c c | c c }
\toprule 

\multicolumn{1}{c|}{\multirow{2}{*}{}}

&\multicolumn{4}{c|}{{Gender}}  
&\multicolumn{4}{c|}{{Race}}  
&\multicolumn{4}{c|}{{Gender $\times$ Race}} 
&\multicolumn{4}{c}{{Gender $\times$ Race $\times$ Age}} 
\\
\cmidrule(lr){2-5} \cmidrule(lr){6-9} \cmidrule(lr){10-13} \cmidrule(lr){14-17}

&\multicolumn{2}{c|}{{$\mathcal{T}_\text{basic}$}}  
&\multicolumn{2}{c|}{{$\mathcal{T}_\text{complex}$}}  
&\multicolumn{2}{c|}{{$\mathcal{T}_\text{basic}$}}  
&\multicolumn{2}{c|}{{$\mathcal{T}_\text{complex}$}}  
&\multicolumn{2}{c|}{{$\mathcal{T}_\text{basic}$}}  
&\multicolumn{2}{c|}{{$\mathcal{T}_\text{complex}$}}  
&\multicolumn{2}{c|}{{$\mathcal{T}_\text{basic}$}}  
&\multicolumn{2}{c}{{$\mathcal{T}_\text{complex}$}}  
\\
\cmidrule(lr){2-3} \cmidrule(lr){4-5}  
\cmidrule(lr){6-7} \cmidrule(lr){8-9} 
\cmidrule(lr){10-11} \cmidrule(lr){12-13}
\cmidrule(lr){14-15} \cmidrule(lr){16-17}

Occupation
& SD
& EFA
& SD
& EFA
& SD
& EFA
& SD
& EFA
& SD
& EFA
& SD
& EFA
& SD
& EFA
& SD
& EFA
\\
\midrule

Analyst & 0.68 & 0.00 & 0.74 & 0.03 & 0.48 & 0.02 & 0.95 & 0.04 & 0.44 & 0.04 & 0.81 & 0.07 & 0.65 & 0.04 & 0.46 & 0.10 \\
Assistant & 1.00 & 0.04 & 0.93 & 0.01 & 0.82 & 0.05 & 0.66 & 0.03 & 0.84 & 0.02 & 0.69 & 0.09 & 0.27 & 0.04 & 0.25 & 0.06 \\
Attendant & 0.38 & 0.01 & 0.35 & 0.04 & 0.55 & 0.04 & 0.53 & 0.11 & 0.38 & 0.04 & 0.32 & 0.01 & 0.41 & 0.08 & 0.23 & 0.12 \\
Baker & 1.00 & 0.01 & 0.95 & 0.04 & 0.91 & 0.01 & 0.38 & 0.09 & 0.92 & 0.02 & 0.46 & 0.03 & 0.78 & 0.03 & 0.53 & 0.07 \\
CEO & 0.88 & 0.01 & 1.00 & 0.07 & 0.41 & 0.02 & 0.51 & 0.05 & 0.46 & 0.02 & 0.58 & 0.06 & 0.34 & 0.05 & 0.35 & 0.07 \\
Carpenter & 0.96 & 0.01 & 0.90 & 0.05 & 0.62 & 0.05 & 0.53 & 0.09 & 0.66 & 0.04 & 0.56 & 0.04 & 0.76 & 0.02 & 0.80 & 0.07 \\
Cashier & 0.85 & 0.00 & 0.53 & 0.01 & 0.43 & 0.02 & 0.84 & 0.04 & 0.44 & 0.04 & 0.62 & 0.02 & 0.33 & 0.09 & 0.61 & 0.09 \\
Cleaner & 0.23 & 0.14 & 0.99 & 0.19 & 0.48 & 0.02 & 0.90 & 0.08 & 0.34 & 0.01 & 0.91 & 0.08 & 0.18 & 0.05 & 0.32 & 0.06 \\
Clerk & 0.99 & 0.00 & 0.49 & 0.01 & 0.37 & 0.05 & 0.41 & 0.02 & 0.46 & 0.02 & 0.41 & 0.07 & 0.14 & 0.02 & 0.25 & 0.04 \\
Construction worker & 0.94 & 0.16 & 0.59 & 0.16 & 0.86 & 0.04 & 0.63 & 0.10 & 0.86 & 0.01 & 0.52 & 0.04 & 0.39 & 0.06 & 0.29 & 0.09 \\
Cook & 0.01 & 0.07 & 0.65 & 0.04 & 0.25 & 0.06 & 0.47 & 0.02 & 0.14 & 0.01 & 0.44 & 0.04 & 0.44 & 0.03 & 0.47 & 0.08 \\
Counselor & 0.66 & 0.01 & 0.25 & 0.00 & 0.42 & 0.05 & 0.84 & 0.08 & 0.37 & 0.04 & 0.51 & 0.04 & 0.31 & 0.03 & 0.54 & 0.09 \\
Designer & 0.98 & 0.10 & 0.85 & 0.00 & 0.35 & 0.02 & 0.29 & 0.03 & 0.43 & 0.03 & 0.38 & 0.04 & 0.22 & 0.05 & 0.25 & 0.05 \\
Developer & 0.66 & 0.03 & 0.25 & 0.13 & 0.74 & 0.02 & 0.40 & 0.08 & 0.64 & 0.01 & 0.26 & 0.06 & 0.53 & 0.05 & 0.29 & 0.17 \\
Doctor & 0.88 & 0.01 & 0.35 & 0.05 & 0.85 & 0.02 & 0.38 & 0.04 & 0.81 & 0.03 & 0.24 & 0.08 & 0.41 & 0.04 & 0.44 & 0.03 \\
Driver & 0.29 & 0.01 & 1.00 & 0.00 & 0.73 & 0.02 & 0.68 & 0.02 & 0.51 & 0.01 & 0.73 & 0.03 & 0.47 & 0.03 & 0.48 & 0.05 \\
Farmer & 0.90 & 0.10 & 0.15 & 0.19 & 0.58 & 0.02 & 0.27 & 0.06 & 0.61 & 0.04 & 0.16 & 0.02 & 0.85 & 0.04 & 0.81 & 0.05 \\
Guard & 0.16 & 0.41 & 0.89 & 0.38 & 0.55 & 0.05 & 0.64 & 0.07 & 0.33 & 0.01 & 0.66 & 0.07 & 0.32 & 0.06 & 0.25 & 0.09 \\
Hairdresser & 0.45 & 0.05 & 1.00 & 0.07 & 0.74 & 0.08 & 0.27 & 0.08 & 0.59 & 0.03 & 0.33 & 0.01 & 0.51 & 0.19 & 0.48 & 0.20 \\
Housekeeper & 0.64 & 0.13 & 0.94 & 0.06 & 0.85 & 0.04 & 0.66 & 0.05 & 0.69 & 0.04 & 0.69 & 0.07 & 0.34 & 0.03 & 0.65 & 0.06 \\
Janitor & 0.88 & 0.19 & 0.13 & 0.13 & 0.76 & 0.02 & 0.83 & 0.06 & 0.75 & 0.04 & 0.41 & 0.02 & 0.32 & 0.07 & 0.29 & 0.06 \\
Laborer & 0.28 & 0.13 & 0.93 & 0.11 & 0.21 & 0.07 & 0.27 & 0.15 & 0.16 & 0.04 & 0.34 & 0.03 & 0.27 & 0.05 & 0.61 & 0.09 \\
Lawyer & 0.95 & 0.05 & 0.88 & 0.01 & 0.98 & 0.02 & 0.89 & 0.06 & 0.95 & 0.02 & 0.84 & 0.02 & 0.48 & 0.04 & 0.42 & 0.06 \\
Librarian & 0.93 & 0.05 & 0.99 & 0.01 & 0.38 & 0.04 & 0.33 & 0.02 & 0.44 & 0.04 & 0.39 & 0.02 & 0.69 & 0.05 & 0.75 & 0.05 \\
Manager & 0.60 & 0.00 & 0.09 & 0.03 & 0.66 & 0.03 & 0.33 & 0.02 & 0.60 & 0.01 & 0.16 & 0.04 & 0.38 & 0.05 & 0.38 & 0.09 \\
Mechanic & 0.99 & 0.04 & 0.28 & 0.07 & 0.78 & 0.05 & 0.43 & 0.16 & 0.80 & 0.01 & 0.25 & 0.03 & 0.41 & 0.05 & 0.42 & 0.08 \\
Nurse & 0.99 & 0.16 & 0.95 & 0.01 & 0.32 & 0.03 & 0.66 & 0.02 & 0.41 & 0.01 & 0.69 & 0.06 & 0.79 & 0.05 & 0.71 & 0.07 \\
Physician & 0.74 & 0.05 & 0.99 & 0.01 & 0.71 & 0.06 & 0.33 & 0.04 & 0.66 & 0.02 & 0.36 & 0.04 & 0.39 & 0.03 & 0.32 & 0.03 \\
Receptionist & 1.00 & 0.03 & 1.00 & 0.06 & 0.38 & 0.09 & 0.30 & 0.08 & 0.46 & 0.08 & 0.40 & 0.05 & 0.81 & 0.11 & 0.75 & 0.13 \\
Salesperson & 0.28 & 0.01 & 0.49 & 0.01 & 0.83 & 0.01 & 0.42 & 0.02 & 0.51 & 0.04 & 0.36 & 0.06 & 0.63 & 0.05 & 0.27 & 0.12 \\
Secretary & 0.98 & 0.03 & 1.00 & 0.00 & 0.98 & 0.08 & 0.94 & 0.05 & 0.96 & 0.01 & 0.95 & 0.07 & 0.43 & 0.07 & 0.20 & 0.05 \\
Sheriff & 0.32 & 0.06 & 0.94 & 0.04 & 0.30 & 0.02 & 0.59 & 0.04 & 0.21 & 0.02 & 0.63 & 0.04 & 0.80 & 0.02 & 0.89 & 0.07 \\
Supervisor & 0.95 & 0.01 & 1.00 & 0.15 & 0.21 & 0.02 & 0.95 & 0.06 & 0.28 & 0.02 & 0.96 & 0.05 & 0.21 & 0.05 & 0.27 & 0.07 \\
Tailor & 0.41 & 0.01 & 0.99 & 0.00 & 0.61 & 0.02 & 0.17 & 0.13 & 0.46 & 0.01 & 0.29 & 0.03 & 0.24 & 0.04 & 0.40 & 0.06 \\
Teacher & 0.91 & 0.03 & 0.80 & 0.04 & 0.99 & 0.03 & 0.71 & 0.09 & 0.94 & 0.02 & 0.64 & 0.04 & 0.30 & 0.05 & 0.40 & 0.07 \\
Writer & 0.69 & 0.01 & 0.32 & 0.00 & 0.67 & 0.03 & 0.33 & 0.02 & 0.58 & 0.01 & 0.14 & 0.06 & 0.44 & 0.04 & 0.57 & 0.05 \\
\midrule
Average & 0.71 & 0.06 & 0.71 & 0.06 & 0.60 & 0.04 & 0.55 & 0.06 & 0.56 & 0.03 & 0.50 & 0.05 & 0.45 & 0.05 & 0.46 & 0.08 \\
\bottomrule
\end{tabular}
}
\caption{
Deviation ratio per occupation of the original SD and our model.
}
\label{tab:supp_deviation_ratio}
\end{table*}

\end{document}